\documentclass[letterpaper]{article} % DO NOT CHANGE THIS
\usepackage{aaai23}  % DO NOT CHANGE THIS
\usepackage{times}  % DO NOT CHANGE THIS
\usepackage{helvet}  % DO NOT CHANGE THIS
\usepackage{courier}  % DO NOT CHANGE THIS
\usepackage[hyphens]{url}  % DO NOT CHANGE THIS
\usepackage{graphicx} % DO NOT CHANGE THIS
\usepackage{natbib}  % DO NOT CHANGE THIS AND DO NOT ADD ANY OPTIONS TO IT
\usepackage{caption} % DO NOT CHANGE THIS AND DO NOT ADD ANY OPTIONS TO IT
\usepackage[linesnumbered,ruled]{algorithm2e}
\usepackage{newfloat}
\usepackage{listings}
\DeclareCaptionStyle{ruled}{labelfont=normalfont,labelsep=colon,strut=off} % DO NOT CHANGE THIS

\usepackage{rotating}
\usepackage{dsfont}
\usepackage{amsmath}
\usepackage[linesnumbered,ruled]{algorithm2e}

\usepackage{xcolor}
\usepackage{pifont}
\usepackage{amssymb}
\newcommand{\cmark}{\textcolor{green!80!black}{\ding{51}}}
\newcommand{\xmark}{\textcolor{red}{\ding{55}}}
\usepackage{multirow}
\usepackage{setspace}
\title{Heterogeneous Graph Learning for Multi-modal Medical Data Analysis}
\author{
    Sein Kim\textsuperscript{\rm 1},
    Namkyeong Lee\textsuperscript{\rm 1},
    Junseok Lee\textsuperscript{\rm 1},
    Dongmin Hyun\textsuperscript{\rm 2},
    Chanyoung Park\textsuperscript{\rm 1,\rm3}\thanks{Corresponding author}
}
\affiliations{
    \textsuperscript{\rm 1} Dept. of Industrial and Systems Engineering, KAIST, Daejeon, Republic of Korea\\
    \textsuperscript{\rm 2} Institute of Artifical Intelligence, POSTECH, Pohang, Republic of Korea\\
    \textsuperscript{\rm 3} Graduate School of Artificial Intelligence, KAIST, Daejeon, Republic of Korea\\
    rlatpdlsgns@kaist.ac.kr, namkyeong96@kaist.ac.kr, junseoklee@kaist.ac.kr, dm.hyun@postech.ac.kr, cy.park@kaist.ac.kr
}

\usepackage{bibentry}
\begin{document}

\maketitle

\begin{abstract}
% Along with recent success of Deep Convolutional Neural Networks (CNNs) in computer vision domain, CNNs have been successfully applied for medical image analysis, which plays a vital role in patient healthcare due to its versatility.
% On the other hand, routine clinical visits of a patient produce not only image data but also non-image data containing clinical information regarding the patient.
Routine clinical visits of a patient produce not only image data, but also non-image data containing clinical information regarding the patient, i.e., medical data is multi-modal in nature.
Such heterogeneous modalities offer different and complementary perspectives on the same patient, resulting in more accurate clinical decisions when they are properly combined.
However, despite its significance, how to effectively fuse the multi-modal medical data into a unified framework has received relatively little attention.
In this paper, we propose an effective graph-based framework called \textbf{HetMed} (\textbf{Het}erogeneous Graph Learning for Multi-modal \textbf{Med}ical Data Analysis) for fusing the multi-modal medical data.
% into a multiple network.
Specifically, we construct a multiplex network that incorporates multiple types of non-image features of patients to capture the complex relationship between patients in a systematic way, which leads to more accurate clinical decisions.
% By constructing multiple types of edges in a multiplex network, our framework truly captures complex relationship between patients, 
% leading to more accurate clinical decision.
Extensive experiments on various real-world datasets demonstrate the superiority and practicality of {HetMed}. The source code for {HetMed} is available at \url{https://github.com/Sein-Kim/Multimodal-Medical}.
\end{abstract}

\section{Introduction}

% Recently, electronic health record (EHR) systems have been rapidly adopted in medical field leading to explosion in the amount of digital medical information \cite{shickel2017deep}.
% Among the information, medical imaging plays a vital role in patient healthcare due to its versatility (e.g., disease prevention, early detection, diagnosis and treatment) \cite{taleb20203d}.
% As the cornerstone of the recent advances in computer vision, deep convolutional neural networks (CNNs) have been actively applied to medical image analysis \cite{azizi2021big, sun2021context, taleb20203d}.

% However, building a large enough expert annotated medical image with high quality, which are at the core of the success, is time-consuming and costly. In computer vision, recent advances in self-supervised learning, which aims at learning from the supervisory signal from the data itself without any label information, made great strides in situations where only limited label information is available \cite{he2020momentum, chen2020simple, grill2020bootstrap}.
% Although self-supervised methods have also been shown to be effective in medical image analysis \cite{taleb20203d, azizi2021big}, 
{
Along with recent advances of deep convolutional neural networks (CNNs) in computer vision domain, analyzing medical image with CNNs have achieved great success in patient healthcare \cite{azizi2021big, sun2021context, taleb20203d, deng2020deep, aggarwal2021diagnostic}.
% Although many works have achieved great success in patient healthcare using only medical images \cite{deng2020deep, aggarwal2021diagnostic}, 
Despite their success, they pay little attention to the inherent uniqueness of medical data: \textit{medical data is multi-modal in nature}.
}
% they pay little attention to the inherent uniqueness of medical data: \textit{medical data is multi-modal in nature}.
%That is, routine clinical visits of a single patient produce various types of data stream which contains information concerning a patient. Those various types of data stream can be divided into two categories, i.e. image data and non-image data \cite{cui2022deep}.
That is, routine clinical visits of a patient produce not only image data, but also non-image data containing clinical information regarding the patient \cite{cui2022deep}, which offers complementary diagnostic information of patients.
More precisely, image data includes images of various body parts used for diagnostic or treatment purposes, while non-image data includes clinical data (e.g., demographic features and diagnosis) and lab test results (e.g., structured genomic sequences and blood test results).
Such heterogeneous medical data provides different and complementary views of the same patient, leading to more accurate clinical decisions when they are properly combined.
Hence, it is crucial to study how to integrate the multi-modal medical data for medical image analysis, which is relatively under explored despite its importance.

% Integrating the non-image data into decision, a preemptive clinical decision can be made at the early stage of dementia, which may not be detected with unimodal image data.
%Thus it is crucial how to integrate those types of multimodal data for medical image classification. 
% Regarding that medical staffs in real world make clinical decisions based on the multiple sources of data, it is natural to consider multimodalities for medical image classification.
%Since these heterogeneous data offers independent complementary diagnostic value, incorporating both modalities may help the model to learn multiple aspects of a single patient thereby making better clinical decisions. 
%However, how to integrate the multiple sources of medical data for medical image classification has been relatively under explored. 

However, effectively fusing the multi-modal medical data is not a trivial task since a variety of clinical modalities contain their own distinct information and may have different data format \cite{cui2022deep}. 
Some studies reflect multiple modalities in an ``early fusion" manner by combining images of different modalities {(e.g., positron emission tomography (PET), computed tomography (CT) and magnetic resonance imaging (MRI))} before training the model \cite{teramoto2016automated, tan2020multimodal, guo2019deep}.
On the other hand, ``late fusion'' approaches combine representations of different imaging modalities \cite{liu2021thyroid, suk2014hierarchical, xu2016multimodal}, while others learn from both image and non-image medical data by combining information from independently trained models in a post-hoc manner \cite{sanyal2021carcinoma, akselrod2019predicting, cheerla2019deep}.
{A recent approach proposes end-to-end learning strategies for fusing multi-modal features at different stages of the model training, i.e., fusing intermediate features and output probabilities \cite{holste2021end}.}
% A recent approach proposes an end-to-end learning of feature fusion, by learning the representations of multiple modalities simultaneously, which are then concatenated to produce the final prediction \cite{holste2021end}.
However, current practice of naively integrating different modalities cannot fully benefit from the complementary relationship between multiple modalities. 
% We argue that the lack of labeled medical image can be supplemented with the complementary diagnostic value offered by multi-modality of medical data.

% \begin{figure}[t]
% \centering
% \includegraphics[width=0.99\columnwidth]{imgs/fig1.pdf} % Reduce the figure size so that it is slightly narrower than the column. Don't use precise values for figure width.This setup will avoid overfull boxes.
% \caption{Average pairwise cosine similarity of patients' non-image features between various CDR levels of dementia. Similarity is calculated with all features in (a), Type 1 features (i.e., personal and family historical list of patients) in (b), and Type 2 features (i.e., cognitive abilities reported by clinician) in (c).}
% \label{fig1}
% \end{figure}

In this work, we focus on the fact that patients who share similar non-image data are likely to suffer from the same disease. 
For example, it is well known that a group of people possessing the E4 allele of apolipoprotein E (APOE) has the primary genetic risk factor for the sporadic form of Alzheimer’s Disease (AD) \cite{emrani2020apoe4}. 
% \textcolor{red}{This implies that unlabeled medical images can be readily classified with a little help from labeled medical images whose non-image data is similar.}
Hence, by introducing non-image data (e.g., APOE in the tabular data) in addition to image data, we argue that preemptive clinical decisions can be made at an early stage of AD, which may not be detected based on image data only.
% By fusing the non-image data (e.g., APOE in the tabular data), preemptive clinical decisions can be made at an early stage of dementia, which may not be detected based on image data only.
% This implies that considering non-image data facilitates preemptive clinical decisions at an early stage of dementia, which may not be detected based on image data only.

\begin{figure}[t]
\centering
\includegraphics[width=0.99\columnwidth]{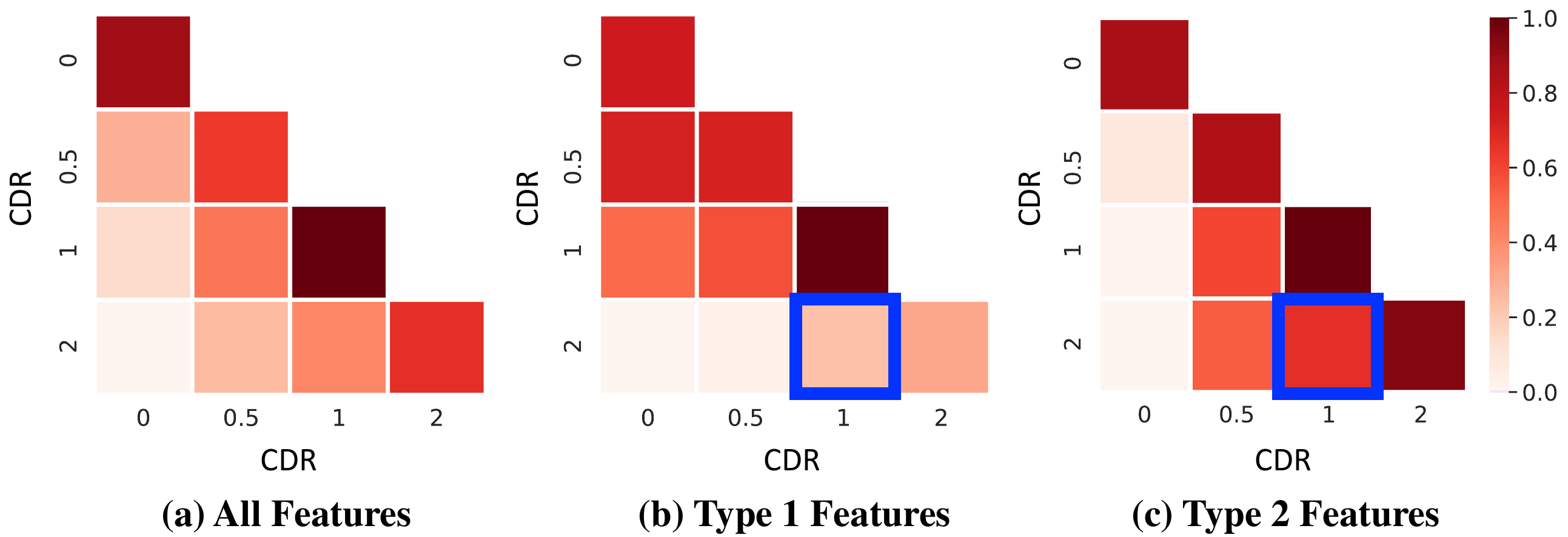} % Reduce the figure size so that it is slightly narrower than the column. Don't use precise values for figure width.This setup will avoid overfull boxes.
\caption{Average pairwise cosine similarity of patients' non-image features between various CDR levels of dementia. Similarity is calculated with all features in (a), Type 1 features (i.e., personal and family historical list of patients) in (b), and Type 2 features (i.e., cognitive abilities reported by clinicians) in (c).}
\vspace{-2ex}
\label{fig1}
\end{figure}

\looseness=-1
To elaborate our argument, we compare the similarity of non-image data between various levels of dementia with OASIS-3 \cite{lamontagne2019oasis} dataset in which patients in the dataset are divided into four classes according to the severity of dementia assessed using Clinical Dementia Rating (CDR) scale \cite{morris1991clinical}: CDR 0 indicates normal cognitive function, CDR 0.5 indicates very mild impairment, CDR 1 indicates mild impairment, and CDR 2 indicates moderate dementia.
% To elaborate our argument, we calculate the average pairwise cosine similarity of patients within a class and between different classes using non-image data provided in OASIS-3 \cite{lamontagne2019oasis} dataset in Figure \ref{fig1} (a).
We calculate the average pairwise cosine similarities of patients' non-image features within a class and between different classes in Figure \ref{fig1} (a).
We observe that the patients in the same class (i.e., diagonal) share high similarity compared with those in different classes (i.e., off-diagonal), demonstrating that patients with similar non-image data are more likely to suffer from the same disease.
Furthermore, we find out that diverse aspects of non-image data induce complex similarity relationships between patients as shown in Figure \ref{fig1} (b) and (c), i.e., the similarity among patients between different classes varies according to which features are selected for calculating the similarity.
As a concrete example, consider two patients each of whom suffers from dementia of CDR 1 and 2, respectively. 
Incorporating personal and family historical list of patients (i.e., Type 1 features) may fail to find connections between the patients of CDR 1 and 2 as shown in Figure \ref{fig1} (b).
% Subset 2 features (i.e., APOE genotype, memory and orientation ability) may fail to find connection between the patients of CDR 1 and 2 as shown in Figue \ref{fig1} (c).
However, if the similarity is calculated based on the patients' cognitive abilities (i.e., Type 2 features), we observe that the similarity between patients that belong to CDR 1 and 2 is relatively high, which indicates that mild impairment (i.e., CDR 1) likely leads to moderate dementia (i.e., CDR 2).
% However, we observe in Figure \ref{fig1} (c) that the patients share high similarity in their cognitive abilities (i.e., Subset 2 features), indicating high probability of mild impairment (i.e., CDR 1) transferring to moderate dementia (i.e., CDR 2).
% However, we observe in Figure \ref{fig1} (c) that the patients exhibit stronger similarity when the similarity is calculated based on a Subset 1 features, i.e., age and judgement ability, indicating high probability of mild impairment (i.e., CDR 1) transferring to moderate dementia (i.e., CDR 2).
This implies that considering complex relationship between patients induced by various types of features helps to capture implicit relationships that can play a key role in making medical decisions.

% As a concrete example, consider two patients each of whom suffers from dementia of CDR 0.5 and 1, respectively. As shown in Figure \ref{fig1} (b), incorporating Subset 1 features, i.e., age and judgement ability, may fail to find the connection between the patients of CDR 0.5 and 1 (i.e., low similarity).
% However, we observe in Figure \ref{fig1} (c) that the patients of CDR 0.5 and 1 exhibit stronger similarity when the similarity is calculated based on a Subset 2 features, i.e., APOE genotype information, memory and orientation ability, which strongly motivates us to regard the complex relationships between the patients.

\smallskip
\looseness=-1
\noindent \textbf{Contribution.}
In this work, we propose a general framework called \textbf{HetMed} (\textbf{Het}erogeneous Graph Learning for Multi-modal \textbf{Med}ical Data Analysis) for fusing multi-modal medical data (i.e., image and non-image) based on a graph structure, which provides a natural way of representing patients and their similarities \cite{parisot2017spectral}.
Specifically, each node in a graph denotes a patient associated with multi-modal features including medical images and non-image data of the patient. 
Moreover, each edge represents the similarity between patients in terms of non-image data.
% With the graph structure, medical images are able to utilize other medical images which may contain common potential causative factors.
% To model complex relationships between patients, we propose three different graph-construction methods as follows: 
% 1) general graph structure whose edges are connected if cosine similarity of patients' non-image data is above threshold,
% 2) fully connected graph whose edge weights indicate level of similarity, 
% 3) multiplex network \cite{de2013mathematical} whose edges are connected according to the similarity of various combinations of features, revealing various types of relationship between patients.
% Our in-depth analyses show that fusing multi-modal medical data based on a graph structure consistently outperforms existing naive fusion methods.
To capture the complex relationship between patients in a systematic way, we propose to construct a multiplex network \cite{de2013mathematical} whose edges are connected according to the similarity of various feature combinations, revealing various types of relationship between patients.
Our extensive experiments on various real-world datasets demonstrate the superiority of HetMed, showing that modeling complex relationships inherent between patients is crucial.
A further appeal of integrating multi-modality into a graph structure, especially via a multiplex network, is that it shows robustness even with scarce label information.

\section{Related Work}
\subsection{Medical Image Representation Learning}
\label{sec:Rel1}
% With the recent advances in computer vision, deep convolutional neural networks (CNNs) have been actively applied to medical image analysis. 
% For medical image classification,
% \citet{li2014medical} introduces customized CNNs with shallow convolutional layer to classify image patches of lung disease, also showing the system can be generalized to other medical image datasets.
% \citet{kermany2018identifying} 
% Based on the ResNet architecture \cite{he2016deep}, U-Net \cite{ronneberger2015u}
% \citet{singla2018subject2vec} proposes an attention-based method that can learn clinically interpretable subject-level representation for predicting disease severity.
% \textcolor{red}{Write more representative works.}
% Most of the prior works in medical image analysis, focus on supervision/self-supervision to get meaningful representation of image. In supervision, based on architecture of CNN, U-Net or ResNet \cite{he2016deep}, achieve dramatic improvement on classification or segmentation tasks. Using CNN architecture and attention mechanism to aggregates local image features to subject representation \cite{singla2018subject2vec}. And also, using graph neural networks, achieves explainable framework for brain network based disease analysis \cite{cui2021brainnnexplainer}.
Training deep neural networks requires massive number of labeled data, which is time-consuming and expensive especially in medical domain.
Recently, self-supervised representation learning methods for medical image has been recently proposed to alleviate the lack of training data.
% However, training the deep neural networks requires massive number of labeled data, which is time-consuming and expensive especially in medical domain.
% Recently, self-supervised representation learning methods for medical image has been recently proposed to alleviate the lack of training data.
Inspired by SimCLR \cite{SimCLR}, \citet{azizi2021big} propose a loss function to maximize the mutual information between the images of the same patient.
\citet{sun2021context} propose bi-level self-supervised learning objective for local anatomical level and patient-level. They use graph structure to model the relationship between different anatomical regions.
For 3D medical images, \citet{taleb20203d} propose five self-supervised learning strategies inspired by recent computer vision approaches \cite{noroozi2016unsupervised, gidaris2018unsupervised, doersch2015unsupervised}.
% Despite the success of CNNs on various tasks of medical image analysis, integrating multi-modality of medical data has been less explored.
Despite their success, they are designed to utilize only single-modality data, whereas multi-modal data are common in medical field.

\subsection{Multi-modal Medical Image Analysis}
\label{sec:Rel2}
By incorporating multiple modalities of medical data, machine learning models can trace patterns of diseases which cannot be captured by single modality of data.
Some studies create multi-modal inputs to CNNs by combining images of multiple modalities (e.g., PET, CT and MRI) in an ``early fusion" manner.
\citet{teramoto2016automated} identify initial pulmonary nodule candidates from both PET and CT images, and candidate regions from two images were combined for classification.
\citet{tan2020multimodal} propose medical image fusion method based on boundary measure modulated by a pulse-coupled neural network \cite{wang2018novel}.
Others learn from both image and non-image medical data by combining information from independently trained models in a post-hoc manner.
\citet{akselrod2019predicting} integrate XGBoost selected clinical features and mammography images to predict breast cancer of patients.
\citet{cheerla2019deep} estimate the future course of patients with cancer lesions by fusing features that are independently learned from clinical data, mRNA expression data, microRNA expression data and histopathology whole slide images (WSIs).
Recently, an end-to-end learning framework for fusing multiple modalities at different stages of model training has been proposed \cite{holste2021end}.
% Recent work \cite{holste2021end} proposes end-to-end learning strategies for fusing multiple modalities at different stages of model training.
However, a naive integration of the modalities cannot fully benefit from the complementary relationship between the modalities.
% Prior works fuse this multi-modality by early fusion or late fusion. In the research of \cite{holste2021end}, suggests three fusion framework: (Probability Fusion, Feature Fusion, Learned Feature Fusion). The probability fusion is one of early fusion method, and the Feature fusion is one of late fusion in this paper. By fusions of multi-modality, this work gets high accuracy on breast cancer classification task.
% However, naively integrating different modalities cannot fully benefit from the complementary relationship between multiple modalities.

To address the issue, some recent works \cite{parisot2017spectral, kazi2019inceptiongcn, cao2021using} suggest to fuse the multi-modalities into a graph structure. 
More precisely, each node feature is associated with a patient's imaging feature vector, while edges represent the similarity between patients. 
These prior works, however, (a) do not consider inherent complex relationship between patients and (b) fail to show generalizability by only applying to a certain disease (e.g., autism spectrum disease and Alzheimer's disease). Instead, we seek a general framework that encodes multi-modality into elaborately constructed network.

\begin{figure}[t]
    \centering
    \includegraphics[width=1.0\columnwidth]{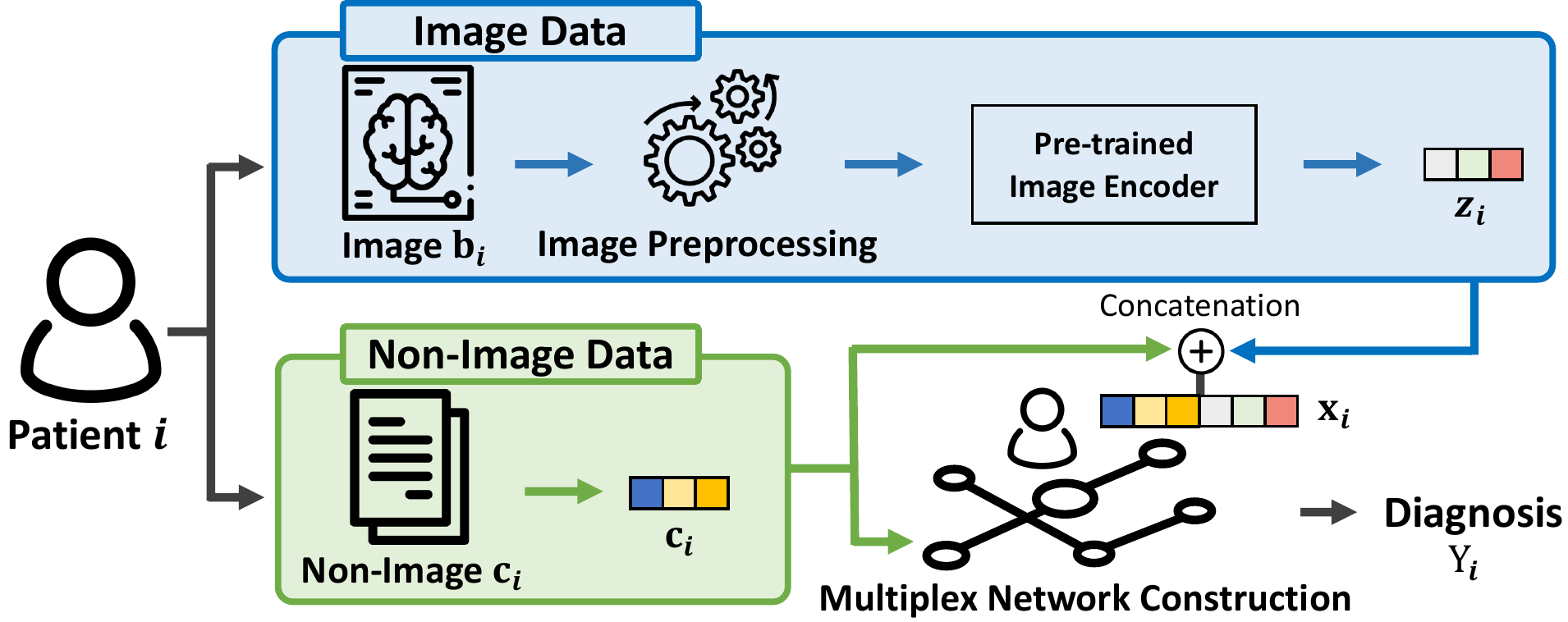}
    % \vspace{-1ex}
    \caption{Overall model framework.}
    \label{fig: framework}
    % \vspace{-2ex}
\end{figure}

\subsection{Multiplex Graph Neural Networks}
\label{sec:Rel3}
With the recent success of deep neural networks, Graph Neural Networks (GNNs) extend the deep neural networks to deal with arbitrary graph-structured data.
{GNNs are trained by repeatedly aggregating the information from neighborhoods in a graph structure \cite{kipf2016semi, hamilton2017inductive, velivckovic2017graph, lee2022grafn}.
Moreover, recent studies try to alleviate lack of label information with self-supervised learning methods \cite{zhu2020deep,lee2022augmentation, lee2022relational}.}
However, by assuming only a single type of relationship between nodes, the above GNNs cannot deal with various types of edges which are prevalent in reality.
{A multiplex network, which is a type of heterogeneous graph \cite{7536145}}, which is also known as multi-view graph \cite{qu2017attention}, multi-layer graph \cite{li2018multi}, multi-dimension graph \cite{ma2018multi}, multi-relational graph \cite{schlichtkrull2018modeling}, and multiplex heterogeneous graph \cite{cen2019representation} in the literature, considers multiple types of relationships among a set of single-typed nodes.
Recent multiplex network embedding methods aim to learn a single embedding for each node that captures multiple types of relationships associated with the node \cite{park2020unsupervised, jing2021hdmi, jing2021graph}.
% By integrating the multiple relationships, recent multiplex network embedding methods aim to learn a single embedding for each node that satisfies all the relationship in the multiplex network \cite{park2020unsupervised, jing2021hdmi, jing2021graph}. 
MVE \cite{qu2017attention} and HAN \cite{wang2019heterogeneous} adopt attention approaches to combine embeddings learned from various relationships. 
DMGI \cite{park2020unsupervised} and HDMI \cite{jing2021hdmi} propose to adopt mutual information-based approaches to learn embeddings of nodes with consensus regularization and high order mutual information, respectively.
To the best of our knowledge, this work is the first to employ a multiplex network to capture the complex relationship between patients for multi-modal medical image analysis.
% Multiplex graph neural networks have been widely used in many applications \cite{du2021new}, such as text summarization \cite{jing2021multiplex}, recommendation system \cite{zhang2020multiplex}, tabular data prediction \cite{guo2021tabgnn} and abstract reasoning \cite{wang2020abstract}.

% Graph is natural architecture to represent relation among subjects by nodes and edges. Especially, graph convolutional networks (GCNs) \cite{kipf2016semi} successes on node classification in semi-supervised method, and also gives motivate for present works about graph. Study about structure of graph, GIN \cite{xu2018powerful} solves graph isomorphic classification with mathematically. In self-supervised learning on graph, DGI \cite{velivckovic2018deep} which maximizes mutual information between global representation of graph and local patch of graph via GCNs.

\section{Problem Statement}
\textbf{Definition 1. (Attributed Multiplex Network)}
An attributed multiplex network is a network $\mathcal{G} = \{\mathcal{G}^{(1)}, ..., \mathcal{G}^{(r)}, ..., \mathcal{G}^{(|\mathcal{R}|)}\} = \{\mathcal{V},\mathcal{E}, \mathbf{X}\}$, where $\mathcal{G}^{(r)} = \{\mathcal{V},\mathcal{E}^{(r)}, \mathbf{X} \}$ is a graph of the relation type $r \in \mathcal{R}$, $\mathcal{V}$ is the set of $n$ nodes, $\mathcal{E} = \bigcup_{r \in \mathcal{R}}{\mathcal{E}^{(r)}}\subseteq \mathcal{V} \times \mathcal{V}$ is the set of all edges with relation type $r$, and $\mathbf{X} \in \mathbb{R}^{|\mathcal{V}| \times F}$ is a matrix that encodes node attribute information for $n$ nodes. Given the network $\mathcal{G}$, $\mathcal{A}=\{\mathbf{A}^{(1)}, ...,  \mathbf{A}^{(|\mathcal{R}|)}\}$ is a set of adjacency matrices, where $\mathbf{A}^{(r)} \in \{ 0, 1\}^{|\mathcal{V}| \times |\mathcal{V}|}$ is an adjacency matrix of the network $\mathcal{G}^{(r)}$.
% Let $\mathcal{G} = ( \mathcal{V},\mathcal{E}, \boldsymbol{\mathrm{X}})$, where $\mathcal{V}$ denotes the set of nodes, $\mathcal{E} \subseteq \mathcal{V} \times \mathcal{V} $ denotes the set of edges. $\mathcal{G}$ is associated with a feature matrix $\mathbf{X} \in \mathbb{R}^{N \times F}$, and an adjacency matrix $\mathbf{A} \in \mathbb{R}^{N \times N}$ where $\mathbf{A}_{ij} = 1$ iff $(v_i, v_j) \in \mathcal{E}$ and $\mathbf{A}_{ij} = 0$ otherwise. 

\smallskip
\noindent\textbf{Task: Multi-modal Medical Image Analysis.}
Given a multi-modal medical data $\mathcal{D} = \{\mathbf{B}, \mathbf{C}, Y\}$, where $\mathbf{B}\in\mathbb{R}^{\mathcal{|V|}\times F_{\text{img}}}$ is the 2D or 3D medical image data, $\mathbf{C}\in\mathbb{R}^{\mathcal{|V|}\times F_{\text{non-img}}}$ is the non-image medical data that consists of categorical and numerical features, and $Y\in\mathbb{R}^{|\mathcal{V}|\times c}$ is the label matrix, where $c$ is the number of classes. The goal of multi-modal medical image analysis is to classify patient $i$ into label $Y_i$ given multi-modal medical data, i.e., medical image $\mathbf{b}_i\in\mathbf{B}\in\mathbb{R}^{\mathcal{|V|}\times F_{\text{img}}}$ and non-image data $\mathbf{c}_i\in\mathbf{C}\in\mathbb{R}^{\mathcal{|V|}\times F_{\text{non-img}}}$ of patient $i$.
% multiple modalities of medical image $x_i$ and non-image data $\mathrm{M}_i$.

\section{Method}

% Our framework fuses multi-modal medical data by constructing multiplex network $\mathcal{G}$ of the patients.
% Each node feature $\mathbf{H}_i$ in the multiplex network contains the information of image data $x_{i}$ and non-image data $\mathrm{M}_{i}$,
% while edges $\mathcal{E}$ indicate the similarity of non-image data $\mathrm{M}_{i}$.
% By dividing non-image information $\mathrm{M}_{i}$ automatically into $|\mathcal{R}|$ subsets, we can finally construct a multiplex network consisting of  $\mathcal{G}^{(r)}$ from each subset $r$.
% Our framework consists of image representation learning, multiplex network construction and training multiplex graph neural network. 
\looseness=-1
An overview of our proposed HetMed is shown in Fig.~\ref{fig: framework}.
% In the following sections, each component of our framework will be detailed.

% Our method aims to construct and learn multiplex network of patient. The node features of graphs consists of image and non-image data of patients. And the edges of multiplex network are constructed by similarities of patients (types of non-image info of patients). We divide types of non-image information automatically in to small subset, and construct graphs from each subset. Our goal is showing multi-modal fusion via multiplex network can get competitive performance and stable from reduction of train set. And also, to show the effect of multi-modal fusion, we fuse non-image information with lots of image encoder from prior works.

% \subsection{Preliminaries}
% To summarize Graph Neural Networks models and introduce our framework, denote s notations for graph and multiplex network in preliminaries. 
\subsection{Image Preprocessing: Anatomical Standardized Image}
Unlike ordinary images, medical images contain noisy information due to the difference in photographing devices and physical size variation per patient. To get medical image represented in standard anatomical information, we use a software called SimpleITK \cite{lowekamp2013design, yaniv2018simpleitk} for medical image pre-processing
that works as:
$\mathbf{b}_{i} = \mathrm{T}_{opt}(\mathrm{T}_{s}(\mathrm{T}_{i}^{-1}(\mathbf{b}_{i}^{original}))$,
% % Medical image contains anatomical information in images unlike ordinary images (i.e. dog, cat, car, etc). 
% % To get image represented in standard anatomical from noisy medical image data due to different photographing device and physical size variation per patients, we use software SimpleITK \cite{lowekamp2013design, yaniv2018simpleitk} for preprocessing of medical images.
% % \begin{equation}
% %     \centering
% %     \mathbf{b}_{i} = \mathrm{T}_{opt}(\mathrm{T}_{s}(\mathrm{T}_{i}^{-1}(\mathbf{b}_{i}^{original})),
% %     \label{eq1}
% % \end{equation}
transforming original image $\mathbf{b}_{i}^{original}$ of patient $i$ into preprocessed image $\mathbf{b}_{i}$ with standard anatomical coordinate. 
Transformation function $\mathrm{T}_{i}^{-1}$ is a fixed mapping function that maps patient $i$'s image domain to patient $i$'s virtual image domain, while $\mathrm{T}_{s}$ is a fixed mapping function that maps virtual image domain to standard anatomical image domain. Finally, $\mathrm{T}_{opt}$ is a modified mapping function for optimization.
The transformation function, $\mathrm{T}_{i}^{-1}$ is a fixed mapping function which maps patient $i$'s image domain to patient $i$'s virtual image domain. $\mathrm{T}_{s}$ is fixed mapping function which maps virtual image domain to standard anatomical image domain. Finally, $\mathrm{T}_{opt}$ is modified mapping function for optimization.

\subsection{Learning Medical Image Representation}
{After the image preprocessing step, we obtain representations of images through a pretrained image encoder.
We adopt several previous self-supervised medical image representation learning methods to verify the generality of HetMed.}
% To verify the generality of our framework, we adopt several previous self-supervised medical image representation learning methods for learning medical image representation.
% In our works, adapt self-supervised image learning models from prior works. In this paper, we focus on the framework of fuse multi-modality. Therefore, we adapt pre-exist image learning model as image encoder. Also, to remove effects of supervision to image encoders before multi-modal fusion, we only select self-supervised learning models.
\subsubsection{4.2.1. 2D Medical Image.}
% \smallskip
% \noindent \textbf{2Dimensional Medical Image Learning.}
% \noindent \textbf{2D Medical Image.}
\looseness=-1
Following \citet{azizi2021big}, we first pre-train an encoder network $f(\cdot):\mathbb{R}^{\mathcal{|V|}\times F_{\text{img}}}\rightarrow\mathbb{R}^{\mathcal{|V|}\times F_z}$ (i.e., a ResNet \cite{he2016deep}) with non-medical image datasets (i.e., STL10 \cite{coates2011analysis} and ImageNet \cite{deng2009imagenet}) by adopting recent self-supervised contrastive learning methods\footnote{We do not use the label information.}, e.g., SimCLR~\cite{SimCLR} and MoCo~\cite{Moco}.
After the non-medical image pre-training step, we adopt Multi-Instance Contrastive Learning (MICLe) loss \cite{azizi2021big}, which maximizes the mutual information between images from the same patient. 
More formally, given a randomly sampled mini-batch of patient $i$, two randomly selected medical images $\mathbf{b}_{i}^{1}$, $\mathbf{b}_{i}^{2}\in\mathbb{R}^{F_{\text{img}}}$ of patient $i$ are encoded via the encoder network $f(\cdot)$ to generate representations $\mathbf{z}_{i}^{1}$, $\mathbf{z}_{i}^{2}\in\mathbb{R}^{F_{\text{z}}}$, respectively. 
Given a mini-batch images of size $N$, the MICLe loss between patient $i$ and other patients is given as follows: 
\begin{equation}
    \mathcal{L}_{i}^{\mathrm{MICLe}} = - \log \frac
    {exp(sim(\mathbf{z}_{i}^{1}, \mathbf{z}_{i}^{2})/\tau)}
    {\sum_{k = 1}^{2N} \mathds{1}_{\left[ k \neq i \right]}exp(sim(\mathbf{z}_{i},\mathbf{z}_{k})/\tau)}
\end{equation}
% $\mathcal{L}_{i}^{\mathrm{MICLe}} = - \log \frac
    % {exp(sim(\mathbf{z}_{i}^{1}, \mathbf{z}_{i}^{2})/\tau)}
    % {\sum_{k = 1}^{2N} \mathds{1}_{\left[ k \neq i \right]}exp(sim(\mathbf{z}_{i},\mathbf{z}_{k})/\tau)},$
% With a $N$ size mini-batch of encoded images, the MICLe loss between patient $i$ and other patients is given as follows:
% For 2D medical image data, we use Multi-Instance Contrastive Loss \cite{azizi2021big}. The MICLe, maximize mutual information between images from same patient.
% \begin{equation}
% \small
%     \centering
%     \mathcal{L}_{i}^{\mathrm{MICLe}} = - \log \frac
%     {exp(sim(\mathbf{z}_{i}^{1}, \mathbf{z}_{i}^{2})/\tau)}
%     {\sum_{k = 1}^{2N} \mathds{1}_{\left[ k \neq i \right]}exp(sim(\mathbf{z}_{i},\mathbf{z}_{k})/\tau)},
%     \label{eq2}
% \end{equation}
where $sim(\cdot, \cdot)$ is the cosine similarity between two vectors, and $\tau$ is a temperature hyperparameter.
Finally, the patient $i$'s image representation $\mathbf{z}_i$ is given by the mean of multiple images, i.e., $\mathbf{z}_i = 1/K\sum_{k = 1}^{K}{\mathbf{z}_i^k}$, where $\mathbf{z}_i^k$ is the representation of the $k$-th image of patient $i$, and $K$ is the number of images for a patient.

% Eq \ref{eq2} is MICLe of patient i's image j. Where, $\mathrm{X}_{i}^{j}$ is patient i's image j, then,  $\mathrm{q}_{i} = Enc(aug(\mathrm{X}_{i}^{j}))$ and $\mathrm{q}'_{i} = Enc(aug(\mathrm{X}_{i}^{l})); \mathrm{X}_{i}^{l} \in \left\{ \mathrm{X}_{i}^{1}, \mathrm{X}_{i}^{2}, \ldots , \mathrm{X}_{i}^{|i|}\right\}$. $Enc$ is image feature extractor of SimCLR or MoCo. For image augmentation (i.e. $aug$), we use resize crop, horizontal flip, gaussian blur. Because all medical image data have one channel, color jitter doesn't use for augmentation function. Like \cite{azizi2021big} did, we load SimCLR model which pretrained with non-medical image data, and train this pretrained model with medical data and MICLe. Not only use SimCLR for image representation extractor, we also use MoCo with MICLe to show stable performance of our works.

% \smallskip
% \noindent \textbf{3D Medical Image.} 
\subsubsection{4.2.2. 3D Medical Image.}
On the other hand, digital medical imaging systems can also create 3D images of human organs. 
With 3D images, medical staffs can access new angles, resolutions, and more details required for better medical decisions while minimizing radiation exposure of patients.
However, inherent limited availability of 3D medical image data \cite{singh20203d} makes it difficult to learn representations of 3D images by directly applying MICLe loss.
% However, Different from 2D images, there are lack of 3D medical images which make difficulty in applying MICLe loss. 
Thus, we adopt recently proposed medical image representation learning approaches for 3D medical image, i.e. 3D Jigsaw, 3D Rotation and 3D Exampler~\cite{taleb20203d} to pre-train an encoder network $f(\cdot)$ that produces a medical image representation $\mathbf{z}_i\in\mathbb{R}^{F_z}$ for each patient $i$. These image representations are later used as node features of the multiplex network. Note that following~\cite{taleb20203d}, we do not pre-train $f(\cdot)$ with non-medical image dataset.
% In 3 Dimensional medical image learning, we adapt 3D - Jigsaw, 3D - Rotation, 3D - Exampler from \cite{taleb20203d}. These three models select as 3 Dimensional image encoder for our framework. Same as 2 Dimensional medical image learning, we use self-supervised model, to remove labels' supervision to image encoder before fuse multi-modality.

% \smallskip
% \noindent\textbf{Discussion. }
% Note that the above techniques for  can be used to pre-train the encoder network, i.e., $f(\cdot)$. The medical image representation $\mathbf{z}_i$ produced by $f(\cdot)$ can used for node features of the multiplex network.
% Note that the above techniques can be used pretrain the image encoder

\begin{figure}[t]
    \centering
    \includegraphics[width=0.99\columnwidth]{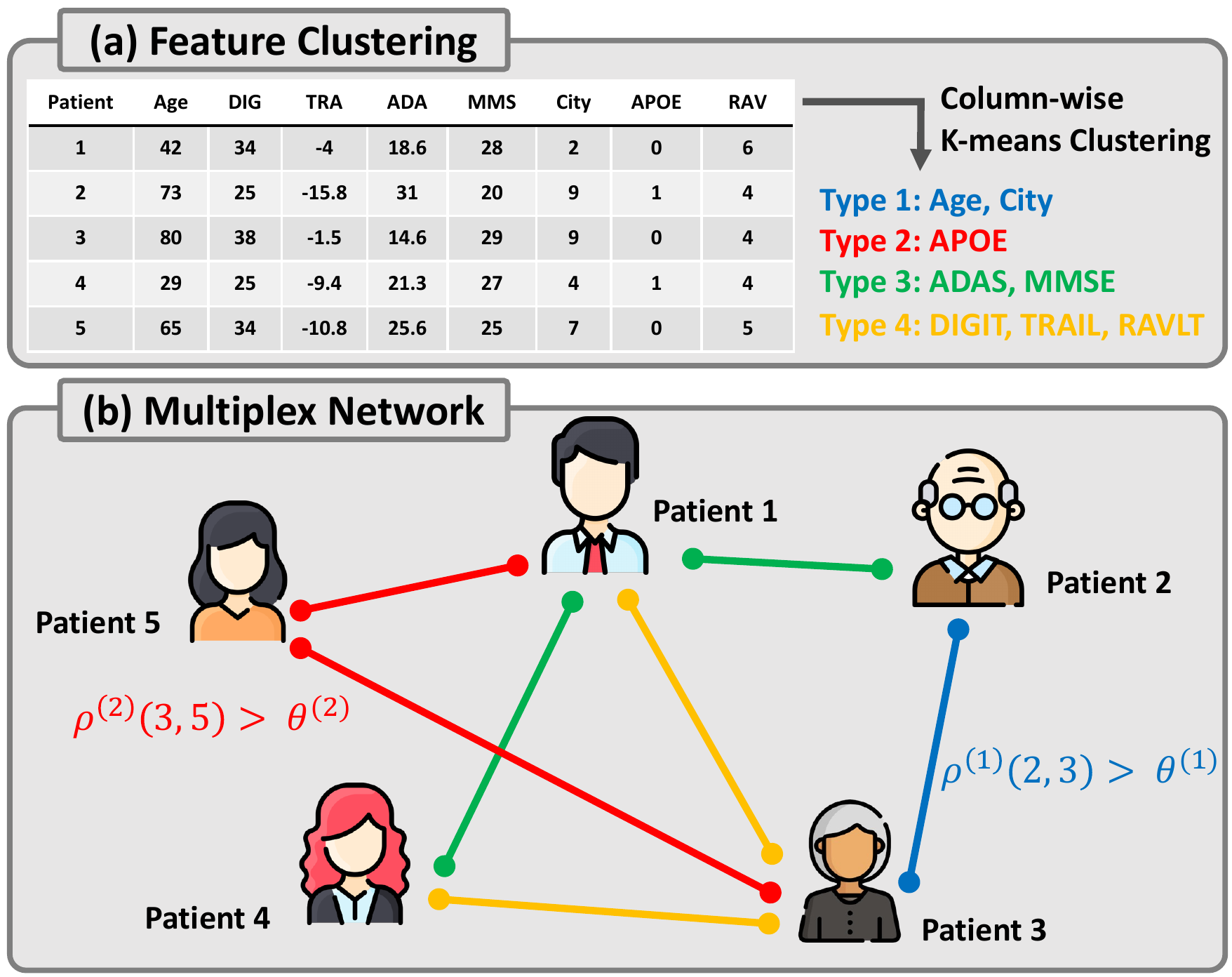}
    % \vspace{-1ex}
    \caption{Multiplex network construction.}
    \label{fig: graph construction}
    % \vspace{-3ex}
\end{figure}

\subsection{Multiplex Network Construction}

In this section, we introduce how to construct a multiplex network to capture the inherent complex relationship between patients. 
Different from conventional approaches that utilize multiplex networks introduced in Section \ref{sec:Rel3}, where multiple types of relations are predefined (e.g., Paper-Author-Paper, Paper-Subject-Paper relationship in citation networks), relationship between patients are usually not given.
Thus, the main challenge of constructing a multiplex network based on patient data is how to extract meaningful relationship between patients.
% Thus, extracting the complex relationship among patients is a key challenge to construct multiplex network.
On the other hand, non-image medical data contains various types of information regarding patients, e.g. demographic features, personal and family historical list and lab test results, each of which provides unique clinical information. 
That is, each type of non-image feature has a different connection to the target disease. 
For example, African Americans (i.e., demographic features) have been reported to have a higher prevalence of Alzheimer's disease than Caucasians \cite{howell2017race}, while historical list (i.e., personal and family historical list) is an important non-modifiable risk factor for hypertension \cite{ranasinghe2015influence}.
Since non-image features can be categorized into several types regarding their characteristics, categorizing the features is the first step for relationship extraction.
% Moreover, since non-image features can be categorized into several types depending on their characteristics, categorizing the non-image features is important.
% Moreover, since variety of non-image features are related to a certain type of features, the features should be categorized to compose heterogeneous types of edges.

% \textcolor{red}{Incorporating multi-modality, there are prevalent method that fuse image data and non-image data. One is concatenate the two modality and the other is construct graph from modalities. Prior works construct single graph to fuse the modalities and also get some positive transfer from structure of graph. But, the features of non-image data have different types. So, constructing single graph from non-image data without considering type of feature has probability that can't contain information from non-image data. Therefore, we separate types of non-image data and build graph for each type. By doing so, we can capture patient relation with multiple aspect.}

Although adopting the domain knowledge is straightforward for the non-image features categorization, domain knowledge may (1) not be always available, and (2) not help discover implicit correlation between features. 
Thus, to automatically categorize various types of non-image features, we simply adopt K-means clustering algorithm~\footnote{For simplicity, we adopt K-means column-wise clustering assuming the categorical features are continuous, because a majority of the categorical features are in fact ordinal features, e.g., 12 out of 14 categorical features, and 13 out of 17 categorical features are ordinal in ADNI and OASIS-3 dataset, respectively. A general approach that would work regardless of the feature type is to use a cluster algorithm that works on mixed data types, such as K-prototype clustering~\cite{huang1997clustering}.} on the non-image tabular data $\mathbf{C} \in \mathbb{R}^{|\mathcal{V}| \times F_{\text{non-img}}}$ in a column-wise manner, which partitions
% Specifically, given the non-image tabular data $\mathbf{C} \in \mathbb{R}^{|\mathcal{V}| \times F_{\text{non-img}}}$, we perform the K-means clustering in the column-wise manner 
% be a non-image tabular dataset where $|V|$ and $F$ denote the number of patients and non-image feature dimensions, respectively. 
% Specifically, let $\mathbf{C} \in \mathbb{R}^{|V| \times F_{c}}$ be a non-image tabular dataset where $|V|$ and $F$ denote the number of patients and non-image feature dimensions, respectively. 
% Then, the K-means clustering algorithm partitions 
the features in $F_{\text{non-img}}$ dimensions into $|\mathcal{R}|$ sets, i.e., $\mathbf{T} = \{ T_1, T_2, \ldots, T_{|\mathcal{R}|} \}$.
% in a column-wise manner.
% by optimizing following objective:
% Start on tabular non-image data set: $\mathrm{C}$ which $\boldsymbol{\mathrm{C}} \in  \mathbb{R}^{\mathrm{N} \times \mathrm{F}}$, where $\mathrm{N}$ is number of patients, $\mathrm{F}$ is number of non-image feature of patient.
% To separate types of non-image feature automatically, we use K-means Clustering. Start on tabular non-image data set: $\mathrm{C}$ which $\boldsymbol{\mathrm{C}} \in  \mathbb{R}^{\mathrm{N} \times \mathrm{F}}$, where $\mathrm{N}$ is number of patients, $\mathrm{F}$ is number of non-image feature of patient.
% \begin{equation}
%     \centering
%     \underset{\mathbf{T}}{\operatorname{argmin}}\sum_{k=1}^{|\mathcal{R}|} \frac{1}{|T_k|}\sum_{\boldsymbol{a},\boldsymbol{b} \in T_k} \begin{Vmatrix} \boldsymbol{a} - \boldsymbol{b} \end{Vmatrix}^{2},
%     \label{eq3}
% \end{equation}
% where $\boldsymbol{a}, \boldsymbol{b} \in \mathbb{R}^{1 \times \mathrm{|V|}}$ are columns of the non-image tabular dataset. 
By doing so, we divide non-image data $\mathbf{C}$ into non-overlapping $|\mathcal{R}|$ types of features as shown in Figure \ref{fig: graph construction} (a).
% With hyperparameter |\mathcal{R}|, by Eq \ref{eq3} we divide non-image data to K number of non-duplicated subsets.

\looseness=-1
With $|\mathcal{R}|$ types of non-image features, we construct a multiplex network $\mathcal{G} = \{\mathcal{G}^{(1)}, ..., \mathcal{G}^{(r)}, ..., \mathcal{G}^{(|\mathcal{R}|)}\}$, where each $\mathcal{G}^{(r)}$ is constructed by calculating the cosine similarity of type $r$ features of non-image data, i.e., $\mathbf{C}^{(r)}$, as follows:
% that consists of graph $\mathcal{G}^{(r)}$ by calculating the cosine similarity of type $r$ features as follows:
\begin{equation}
    \centering
        \mathbf{A}^{(r)}(i, j) = \begin{cases}
                1 &\mbox{if } \rho^{(r)} (i, j) > \theta^{(r)} , \\
                0 &\mbox{otherwise }
                \end{cases}
    \label{eq4}
    % \vspace{-1ex}
\end{equation}
where $\rho^{(r)}(i, j)$ is the cosine similarity of type $r$ non-image feature between patient $i$ and $j$, i.e., $\rho^{(r)} (i, j) =  \frac{\mathbf{c}^{(r)}_i \cdot \mathbf{c}^{(r)}_j}{\| \mathbf{c}^{(r)}_i \| \cdot \| \mathbf{c}^{(r)}_j \|}$ where $\mathbf{c}_i^{(r)}$ is the $i$-th row of $\mathbf{C}^{(r)}$, and $\theta^{(r)}$ is the threshold hyperparameter for each relation $r$.
For example in Figure \ref{fig: graph construction} (b), patient 2 and patient 3 are connected by relation type 1 since their age and city features (i.e. demographics type) are similar, while patient 3 and patient 5 are connected by APOE similarity (i.e., relation type 2).
Finally, the image representation $\mathbf{Z} \in \mathbb{R}^{|\mathcal{V}| \times F_{z}}$,
% , whose $i$-th row denotes the medical image representation $\mathbf{z_i}$ of patient $i$
and non-image feature $\mathbf{C} \in \mathbb{R}^{|\mathcal{V}| \times F_{\text{non-img}}}$ are concatenated to generate the node attribute matrix $\mathbf{X} \in \mathbb{R}^{|\mathcal{V}| \times F}$ of multiplex network $\mathcal{G}$, i.e., $F = (F_{z} + F_{\text{non-img}})$.
% Finally, multiplex network $\mathcal{G}$, whose node attribute feature $\mathbf{X} \in \mathbb{R}^{|V| \times (F_{z} + F_{c})}$ is given as concatenation of learned image representation $\mathbf{Z} \in \mathbb{R}^{|V| \times F_{z}}$ whose row is medical image representation $\mathbf{z_i}$ of patient $i$ and non-image feature $\mathbf{C}$, is constructed.
% Then, multiplex network $\mathcal{G} = $ whose feature $\mathbf{H}$ is given as concatenation of learned image representation $\mathrm{H}$ and non-image feature $\mathrm{M}$ 
% Makes binary adjacency matrix from cosine similarities and threshold  $\theta$. In Eq \ref{eq4}, the cosine similarities of type t non-image features: $    \rho_t (u,v) &=  \frac{M_t(u) \cdot M_t(v)}{\left| M_t(u) \right| \left| M_t(v) \right |} $, where $M_t(u)$ is $\mbox{type t non-image features of } u$, $t \in \left\{T_1, T_2, \ldots, T_{K}\right\}$ By doing so, unlike prior works which use image feature or other features from MRI, CT to cosine similarities form to make adjacency matrix of graph, we can observe aspects and its effects of non-image data only by graphs.

\subsection{Multiplex Graph Neural Networks}
\looseness=-1
Given a multiplex network $\mathcal{G}$, learning from heterogeneous types of relationship among nodes is non-trivial, since these types of relationship are related \cite{park2020unsupervised}.
Recently proposed multiplex network embedding methods captures interaction between relationships through attention mechanism \cite{wang2019heterogeneous} and consensus regularization \cite{park2020unsupervised}.
Among various approaches for multiplex network embedding, we adopt DMGI \cite{park2020unsupervised} in this work due to its simplicity and applicability under unsupervised setting. 
However, as HetMed is model-agnostic, any multiplex network embedding method can be adopted
as will be demonstrated in Appendix \ref{app: Additional Experiments}.
% We adopted DMGI \cite{park2020unsupervised} for medical image analysis.

\smallskip
\noindent \textbf{Deep Multiplex Graph Infomax (DMGI).}
The core idea of DMGI is to learn the consensus embedding of a single node regarding multiple relation types, while each relation-type specific node embedding is trained to maximize the mutual information with relation-type specific summary vector.
Specifically, relation-type specific node encoder $g_r(\cdot): \mathbb{R}^{|\mathcal{V}| \times F} \times \mathbb{R}^{|\mathcal{V}| \times |\mathcal{V}|} \rightarrow \mathbb{R}^{|\mathcal{V}| \times d} $ generates the relation-type specific node embedding matrix $\mathbf{H}^{(r)} \in \mathbb{R}^{|\mathcal{V}| \times d}$ for each relation type $r \in \mathcal{R}$. 
Then, the summary representation $\mathbf{s}^{(r)}$ of graph $\mathcal{G}^{(r)}$ is computed through mean pooling readout function.
Given the embedding $\mathbf{H}^{(r)}$ and the summary vector $\mathbf{s}^{(r)}\in\mathbb{R}^{d}$, DMGI maximizes the mutual information between $\mathbf{H}^{(r)}$ and $\mathbf{s}^{(r)}$, while minimizing the mutual information between corrupted representation $\tilde{\mathbf{H}}^{(r)}$ and $\mathbf{s}^{(r)}$ as
\begin{equation}
\small
    \mathcal{L}^{(r)} = \sum_{i=1}^{\mathcal{|V|}} \log{\mathcal{D}(\mathbf{h}_{i}^{(r)}, \mathbf{s}^{(r)})} +
    \sum_{j = 1}^{\mathcal{|V|}} \log{(1 - \mathcal{D}(\tilde{\mathbf{h}}_{j}^{(r)}, \mathbf{s}^{(r)}))}
\end{equation}
% \begin{equation}
% \small
%     \centering
%     \mathcal{L}^{(r)} = \sum_{i=1}^{\mathcal{|V|}} \log{\mathcal{D}(\mathbf{h}_{i}^{(r)}, \mathbf{s}^{(r)})} +
%     \sum_{j = 1}^{\mathcal{|V|}} \log{(1 - \mathcal{D}(\tilde{\mathbf{h}}_{j}^{(r)}, \mathbf{s}^{(r)}))}
%     \label{eq5}
% \end{equation}
where $\mathcal{D}$ is a discrimination function that scores patch-summary representation pairs, i.e., high score for positive patch-summary pairs $(\mathbf{h}_{i}^{(r)}, \mathbf{s}^{(r)})$, where $\mathbf{h}_{i}^{(r)}$ is the type $r$ embedding of node $v_i$.
However, independently trained encoder $g_r(\cdot)$, which contains relevant information regarding each relation type $r$, cannot fully benefit from the multiplexity of the network.
To this end, DMGI proposes consensus regularization which aims to minimize the discrepancy between relation specific embeddings, i.e., $\{ \mathbf{H}^{(r)} | r \in \mathcal{R} \}$, and the consensus embedding $\mathbf{O} \in \mathbb{R}^{|V| \times d}$ as: 
\begin{equation}
\small
\ell_{cs} = 
    \left[\mathbf{O} - \mathcal{Q}\left(\{\mathbf{H}^{(r)} | r \in \mathcal{R} \}\right) \right]^{2} - 
    \left[\mathbf{O} - \mathcal{Q}\left(\{\tilde{\mathbf{H}}^{(r)} | r \in \mathcal{R} \}\right) \right]^{2}    
\end{equation}
% \begin{equation}
% \small
%     \centering
%     \ell_{cs} = 
%     \left[\mathbf{O} - \mathcal{Q}\left(\{\mathbf{H}^{(r)} | r \in \mathcal{R} \}\right) \right]^{2} - 
%     \left[\mathbf{O} - \mathcal{Q}\left(\{\tilde{\mathbf{H}}^{(r)} | r \in \mathcal{R} \}\right) \right]^{2}
%     \label{eq6}
% \end{equation}
where $\mathcal{Q}$ is an attentive pooling function for every relation-specific embedding matrix $\mathbf{H}^{(r)}$.
Finally, we introduce a semi-supervised module to learn from labeled medical images based on the consensus embedding $\mathbf{O}$ as:
\begin{equation}
  \ell_{sup} = 
    - \frac{1}{|\mathcal{Y}_L|}\sum_{l \in \mathcal{Y}_L}{\sum_{i=1}^{c}{Y_{li}\ln{\hat{Y}_{li}}}}  
\end{equation}
% $\ell_{sup} = 
%     - \frac{1}{|\mathcal{Y}_L|}\sum_{l \in \mathcal{Y}_L}{\sum_{i=1}^{c}{Y_{li}\ln{\hat{Y}_{li}}}}$,
% For medical image classification, we add a semi-supervised loss which predicts the labels of patients based on the consensus embedding $\mathbf{O}$ as follows:
% \begin{equation}
%     \centering
%     \ell_{sup} = 
%     - \frac{1}{|\mathcal{Y}_L|}\sum_{l \in \mathcal{Y}_L}{\sum_{i=1}^{c}{Y_{li}\ln{\hat{Y}_{li}}}}
%     \label{eq7}
% \end{equation}
where $\mathcal{Y}_{L}$ is the set of labeled node indices, $Y \in \mathbb{R}^{|\mathcal{V}| \times c}$ is the label matrix, and $\hat{Y}$ is the model prediction after passing the consensus embedding $\mathbf{O}$ through a softmax layer.
Finally, our model is optimized to minimize the following loss: 
\begin{equation}
   \mathcal{L} = 
    \sum_{r \in \mathcal{R}}{\mathcal{L}^{(r)}} + \alpha \ell_{cs} + \beta \ell_{sup} + \gamma \begin{Vmatrix} \boldsymbol{\Theta} \end{Vmatrix}^{2} 
\end{equation}
% $\mathcal{L} = 
%     \sum_{r \in \mathcal{R}}{\mathcal{L}^{(r)}} + \alpha \ell_{cs} + \beta \ell_{sup} + \gamma \begin{Vmatrix} \boldsymbol{\Theta} \end{Vmatrix}^{2}$,
% \begin{equation}
%     \centering
%     \mathcal{L} = 
%     \sum_{r \in \mathcal{R}}{\mathcal{L}^{(r)}} + \alpha \ell_{cs} + \beta \ell_{sup} + \gamma \begin{Vmatrix} \boldsymbol{\Theta} \end{Vmatrix}^{2}
%     \label{eq7}
% \end{equation}
where $\alpha$, $\beta$, $\gamma$ are adjustable hyperparameters for each loss term, and $\boldsymbol{\Theta}$ is the trainable parameters of our model.

\section{Experiments}
\subsection{Experimental Setup}

\noindent \textbf{Datasets.}
To evaluate our proposed HetMed, we conduct experiments on five multi-modal medical datasets.
Specifically, we use three brain related datasets, and two breast-related datasets.
% Specifically, we use three brain related datasets, i.e., Alzheimer's Disase Neuroimaging Intitiative (ADNI) \cite{petersen2010alzheimer}, Open Access Series of Imaging Studies (OASIS-3) \cite{lamontagne2019oasis} and Autism Brain Imaging Data Exchange (ABIDE) \cite{di2014autism, di2017enhancing}, and two breast-related datasets, i.e., Quantitative Imaging Network Breast (Duke-Breast) \cite{clark2013cancer, Li_Abramson_Arlinghaus_Chakravarthy_Abramson_Sanders_Yankeelov_2016, li2015multiparametric} and Chines Mammography Dataset (CMMD) \cite{cai2019breast, wang2016discrimination}. 
Note that since 3D images can be readily converted to 2D images through slicing, we also report the performance on 3D image datasets when they are converted to 2D.
% for all the evaluations on 2D medical image analysis, we also use 3D images to 2D
The detailed statistics are summarized in Table \ref{tab: data statistics} and further details on each dataset are described in Appendix \ref{app: Datasets}.
\begin{table}[t]
    \centering
    % \footnotesize
    % \renewcommand{\arraystretch}{0.9}
    \resizebox{0.99\columnwidth}{!}{
    \begin{tabular}{c|ccccccc}
    % \multirow{2}{*}{Dataset} & Body & Target & \multirow{2}{*}{3D} & \# Non-Img & \# Sub- & \# Cla- \\
    \multirow{2}{*}{Dataset} & Body & Target & \multirow{2}{*}{3D} & \# Non-Img & \multirow{2}{*}{\# Subjects} & \multirow{2}{*}{\# Classes} \\
    & Parts & Disease & & Features &  & \\
    \hline \hline
    ADNI       & Brain  & Alzheimer & \cmark & 17 & 417 & 3 \\
    OASIS-3    & Brain  & Alzheimer & \cmark & 19 & 979 & 4 \\
    ABIDE      & Brain  & Autism    & \cmark & 14 & 977 & 2 \\
    Duke-Breast & Breast & Tumor     & \xmark & 25 & 614 & 3 \\
    CMMD       & Breast & Tumor     & \xmark & 4 & 1774 & 2 \\ \hline
    \end{tabular}
    }
    % \vspace{-1ex}
\caption{Data statistics.}
\label{tab: data statistics}
% \vspace{-1ex}
\end{table}

\begin{table*}[t]
    % \begin{minipage}{0.65\linewidth}{
	\centering
	\small
	% \renewcommand{\arraystretch}{0.8}
	% \resizebox{1.0\linewidth}{!}{
    \begin{tabular}{c|cc|cc|cc|cc|cc} \multicolumn{1}{c|}{\multirow{2}{*}{Model}}& \multicolumn{2}{c|}{ADNI} & \multicolumn{2}{c|}{OASIS-3} & \multicolumn{2}{c|}{ABIDE} & \multicolumn{2}{c|}{Duke-Breast} & \multicolumn{2}{c}{CMMD}
        \\ \cline{2-11}
        & \multicolumn{1}{c}{Ma-F1} & \multicolumn{1}{c|}{Mi-F1} & \multicolumn{1}{c}{Ma-F1} & \multicolumn{1}{c|}{Mi-F1} & \multicolumn{1}{c}{Ma-F1} & \multicolumn{1}{c|}{Mi-F1} & \multicolumn{1}{c}{Ma-F1} & \multicolumn{1}{c|}{Mi-F1} & \multicolumn{1}{c}{Ma-F1} & \multicolumn{1}{c}{Mi-F1} 
        \\ \hline \hline
        \multirow{2}{*}{Feat. (=MLP)} &
        \multicolumn{1}{c}{0.521} & 0.562 &
        \multicolumn{1}{c}{0.216} & 0.625 &
        \multicolumn{1}{c}{0.407} & 0.753 & 
        \multicolumn{1}{c}{0.466} & 0.712 & 
        \multicolumn{1}{c}{0.667} & 0.75
        \\  
        &
        \multicolumn{1}{c}{\scriptsize{(0.017)}} & \scriptsize{(0.020)} &
        \multicolumn{1}{c}{\scriptsize{(0.008)}} & \scriptsize{(0.022)} &
        \multicolumn{1}{c}{\scriptsize{(0.015)}} & \scriptsize{(0.020)} & 
        \multicolumn{1}{c}{\scriptsize{(0.031)}} & \scriptsize{(0.032)} & 
        \multicolumn{1}{c}{\scriptsize{(0.021)}} & \scriptsize{(0.017)}
        \\
        \multirow{2}{*}{Prob.}  & 
        \multicolumn{1}{c}{0.509} & 0.522 &
        \multicolumn{1}{c}{\textbf{0.230}} & 0.639 & 
        \multicolumn{1}{c}{0.385} & 0.761 & 
        \multicolumn{1}{c}{0.407} & 0.688 & 
        \multicolumn{1}{c}{0.564} & 0.665
        \\
        & 
        \multicolumn{1}{c}{\scriptsize{(0.009)}} & \scriptsize{(0.020)} &
        \multicolumn{1}{c}{\scriptsize{(0.022)}} & \scriptsize{(0.025)} & 
        \multicolumn{1}{c}{\scriptsize{(0.034)}} & \scriptsize{(0.031)} & 
        \multicolumn{1}{c}{\scriptsize{(0.028)}} & \scriptsize{(0.025)} & 
        \multicolumn{1}{c}{\scriptsize{(0.013)}} & \scriptsize{(0.026)}
        \\ 
        \multirow{2}{*}{Learned} &
        \multicolumn{1}{c}{0.576} & 0.598 &
        \multicolumn{1}{c}{0.199} & 0.647 &
        \multicolumn{1}{c}{0.649} & 0.776 & 
        \multicolumn{1}{c}{\textbf{0.484}} & 0.625 & 
        \multicolumn{1}{c}{0.714} & 0.773
        \\
        &
        \multicolumn{1}{c}{\scriptsize{(0.012)}} & \scriptsize{(0.022)} &
        \multicolumn{1}{c}{\scriptsize{(0.009)}} & \scriptsize{(0.010)} &
        \multicolumn{1}{c}{\scriptsize{(0.021)}} & \scriptsize{(0.023)} & 
        \multicolumn{1}{c}{\scriptsize{(0.022)}} & \scriptsize{(0.024)} & 
        \multicolumn{1}{c}{\scriptsize{(0.032)}} & \scriptsize{(0.020)}
        \\\hline
        \multirow{2}{*}{Spec.} &
        \multicolumn{1}{c}{0.628} & 0.788 &
        \multicolumn{1}{c}{0.202} & 0.679 &
        \multicolumn{1}{c}{0.696} & 0.717 & 
        \multicolumn{1}{c}{0.427} & 0.701 & 
        \multicolumn{1}{c}{0.681} & 0.742
        \\
        &
        \multicolumn{1}{c}{\scriptsize{(0.009)}} & \scriptsize{(0.018)} &
        \multicolumn{1}{c}{\scriptsize{(0.011)}} & \scriptsize{(0.014)} &
        \multicolumn{1}{c}{\scriptsize{(0.032)}} & \scriptsize{(0.037)} & 
        \multicolumn{1}{c}{\scriptsize{(0.010)}} & \scriptsize{(0.025)} & 
        \multicolumn{1}{c}{\scriptsize{(0.036)}} & \scriptsize{(0.024)}
        \\
        \multirow{2}{*}{GCN} &
        \multicolumn{1}{c}{0.606} & 0.795&
        \multicolumn{1}{c}{0.201} & 0.670 &
        \multicolumn{1}{c}{0.768} & 0.770 & 
        \multicolumn{1}{c}{0.430} & 0.671 & 
        \multicolumn{1}{c}{0.683} & 0.745
        \\
        &
        \multicolumn{1}{c}{\scriptsize{(0.022)}} & \scriptsize{(0.015)} &
        \multicolumn{1}{c}{\scriptsize{(0.021)}} & \scriptsize{(0.033)} &
        \multicolumn{1}{c}{\scriptsize{(0.019)}} & \scriptsize{(0.017)} & 
        \multicolumn{1}{c}{\scriptsize{(0.016)}} & \scriptsize{(0.020)} & 
        \multicolumn{1}{c}{\scriptsize{(0.028)}} & \scriptsize{(0.016)}
        \\
        \multirow{2}{*}{HetMed} &
        \multicolumn{1}{c}{\textbf{0.774}} & \textbf{0.813} &
        \multicolumn{1}{c}{0.205} & \textbf{0.697} &
        \multicolumn{1}{c}{\textbf{0.778}} & \textbf{0.784} & 
        \multicolumn{1}{c}{0.432} & \textbf{0.794} & 
        \multicolumn{1}{c}{\textbf{0.716}} & \textbf{0.785}
        \\
        &
        \multicolumn{1}{c}{\scriptsize{(0.037)}} & \scriptsize{(0.024)} &
        \multicolumn{1}{c}{\scriptsize{(0.005)}} & \scriptsize{(0.011)} &
        \multicolumn{1}{c}{\scriptsize{(0.035)}} & \scriptsize{(0.033)} & 
        \multicolumn{1}{c}{\scriptsize{(0.024)}} & \scriptsize{(0.057)} & 
        \multicolumn{1}{c}{\scriptsize{(0.008)}} & \scriptsize{(0.010)}
        \\ \cline{1-11}
    \end{tabular}
    % }
    % \vspace{-1ex}
\caption{Performance under end-to-end framework on 2D medical image analysis.}
\label{tab:main_result1}
\end{table*}

\begin{table*}[t]
    % \begin{minipage}{0.65\linewidth}{
	\centering
	\small
	% \renewcommand{\arraystretch}{0.8}
	% \resizebox{1.0\linewidth}{!}{
        \begin{tabular}{c|c|cc|cc|cc|cc|cc}
        \multicolumn{1}{c|}{\multirow{2}{*}{Pretrain}} & \multicolumn{1}{c|}{\multirow{2}{*}{Model}}& \multicolumn{2}{c|}{ADNI} & \multicolumn{2}{c|}{OASIS-3} & \multicolumn{2}{c|}{ABIDE} & \multicolumn{2}{c|}{Duke-Breast} & \multicolumn{2}{c}{CMMD}
        \\ \cline{3-12}
        &  & \multicolumn{1}{c}{Ma-F1} & \multicolumn{1}{c|}{Mi-F1} & \multicolumn{1}{c}{Ma-F1} & \multicolumn{1}{c|}{Mi-F1} & \multicolumn{1}{c}{Ma-F1} & \multicolumn{1}{c|}{Mi-F1} & \multicolumn{1}{c}{Ma-F1} & \multicolumn{1}{c|}{Mi-F1} & \multicolumn{1}{c}{Ma-F1} & \multicolumn{1}{c}{Mi-F1} 
        \\ \hline \hline
        \multicolumn{1}{c|}{\multirow{6}{*}{SimCLR}} & \multirow{2}{*}{MLP} &
        \multicolumn{1}{c}{0.561} & 0.781 &
        \multicolumn{1}{c}{0.219} & 0.646 &
        \multicolumn{1}{c}{0.703} & 0.735 & 
        \multicolumn{1}{c}{0.430} & 0.652 & 
        \multicolumn{1}{c}{0.523} & 0.751
        \\
        \multicolumn{1}{c|}{} & &
        \multicolumn{1}{c} {\scriptsize{(0.022)}} & \scriptsize{(0.031)} &
        \multicolumn{1}{c} {\scriptsize{(0.017)}} & \scriptsize{(0.016)} &
        \multicolumn{1}{c} {\scriptsize{(0.023)}} & \scriptsize{(0.022)} & 
        \multicolumn{1}{c} {\scriptsize{(0.019)}} & \scriptsize{(0.019)} & 
        \multicolumn{1}{c} {\scriptsize{(0.027)}} & \scriptsize{(0.014)}
        \\
        \multicolumn{1}{c|}{} & \multirow{2}{*}{GCN} &
        \multicolumn{1}{c}{0.611} & 0.816 &
        \multicolumn{1}{c}{\textbf{0.235}} & 0.685 &
        \multicolumn{1}{c}{0.751} & 0.756 & 
        \multicolumn{1}{c}{0.440} & 0.698 & 
        \multicolumn{1}{c}{0.625} & 0.745
        \\
        \multicolumn{1}{c|}{} &  &
        \multicolumn{1}{c}{\scriptsize{(0.019)}} & \scriptsize{(0.014)} &
        \multicolumn{1}{c}{\scriptsize{(0.025)}}  & \scriptsize{(0.026)} &
        \multicolumn{1}{c}{\scriptsize{(0.016)}} & \scriptsize{(0.018)} & 
        \multicolumn{1}{c}{\scriptsize{(0.017)}} & \scriptsize{(0.019)} & 
        \multicolumn{1}{c}{\scriptsize{(0.014)}} & \scriptsize{(0.022)}
        \\
        \multicolumn{1}{c|}{} & \multirow{2}{*}{HetMed} & 
        \multicolumn{1}{c}{\textbf{0.851}} & \textbf{0.857} &
        \multicolumn{1}{c}{\textbf{0.235}} & \textbf{0.686} &
        \multicolumn{1}{c}{\textbf{0.833}} & \textbf{0.842} &
        \multicolumn{1}{c}{\textbf{0.447}} & \textbf{0.765} & 
        \multicolumn{1}{c}{\textbf{0.720}} & \textbf{0.781}
        \\
        \multicolumn{1}{c|}{} &  & 
        \multicolumn{1}{c}{\scriptsize{(0.009)}} & \scriptsize{(0.012)} &
        \multicolumn{1}{c}{\scriptsize{(0.020)}} & \scriptsize{(0.017)} &
        \multicolumn{1}{c}{\scriptsize{(0.005)}} & \scriptsize{(0.004)} &
        \multicolumn{1}{c}{\scriptsize{(0.011)}} & \scriptsize{(0.021)} & 
        \multicolumn{1}{c}{\scriptsize{(0.025)}} & \scriptsize{(0.022)}
        \\ \cline{1-12}
        
        \multicolumn{1}{c|}{\multirow{6}{*}{MoCo}} & \multirow{2}{*}{MLP} &
        \multicolumn{1}{c}{0.547} & 0.757 &
        \multicolumn{1}{c}{\textbf{0.247}} & 0.669 &
        \multicolumn{1}{c}{0.708} & 0.739 & 
        \multicolumn{1}{c}{0.439} & 0.699 & 
        \multicolumn{1}{c}{0.531} & 0.748
        \\
        \multicolumn{1}{c|}{} &  & 
        \multicolumn{1}{c}{\scriptsize{(0.012)}} & \scriptsize{(0.020)} &
        \multicolumn{1}{c}{\scriptsize{(0.014)}} & \scriptsize{(0.013)} &
        \multicolumn{1}{c}{\scriptsize{(0.012)}} & \scriptsize{(0.014)} &
        \multicolumn{1}{c}{\scriptsize{(0.020)}} & \scriptsize{(0.027)} & 
        \multicolumn{1}{c}{\scriptsize{(0.033)}} & \scriptsize{(0.021)} 
        \\
        \multicolumn{1}{c|}{} & \multirow{2}{*}{GCN} &
        \multicolumn{1}{c}{0.616} & 0.825 &
        \multicolumn{1}{c}{0.238} & 0.679 &
        \multicolumn{1}{c}{0.734} & 0.749 & 
        \multicolumn{1}{c}{0.445} & 0.716 & 
        \multicolumn{1}{c}{0.611} & 0.752
        \\
        \multicolumn{1}{c|}{} &  &
        \multicolumn{1}{c}{\scriptsize{(0.016)}} & \scriptsize{(0.018)} &
        \multicolumn{1}{c}{\scriptsize{(0.013)}} & \scriptsize{(0.021)} &
        \multicolumn{1}{c}{\scriptsize{(0.022)}} & \scriptsize{(0.030)} & 
        \multicolumn{1}{c}{\scriptsize{(0.026)}} & \scriptsize{(0.024)} & 
        \multicolumn{1}{c}{\scriptsize{(0.019)}} & \scriptsize{(0.021)}
        \\  
        \multicolumn{1}{c|}{} & \multirow{2}{*}{HetMed} & 
        \multicolumn{1}{c}{\textbf{0.832}} & \textbf{0.842} &
        \multicolumn{1}{c}{0.242} & \textbf{0.690} &
        \multicolumn{1}{c}{\textbf{0.855}} & \textbf{0.858} &
        \multicolumn{1}{c}{\textbf{0.446}} & \textbf{0.753} & 
        \multicolumn{1}{c}{\textbf{0.706}} & \textbf{0.764}
        \\  
        \multicolumn{1}{c|}{} & & 
        \multicolumn{1}{c}{\scriptsize{(0.011)}} & \scriptsize{(0.020)} &
        \multicolumn{1}{c}{\scriptsize{(0.030)}} & \scriptsize{(0.022)} &
        \multicolumn{1}{c}{\scriptsize{(0.006)}} & \scriptsize{(0.006)} &
        \multicolumn{1}{c}{\scriptsize{(0.011)}} & \scriptsize{(0.027)} & 
        \multicolumn{1}{c}{\scriptsize{(0.018)}} & \scriptsize{(0.024)}
        \\ \hline
    \end{tabular}
    % }
    % \vspace{-1ex}
    \caption{Performance over pretraining strategies on 2D medical image analysis.}
    \label{tab:main_result2}
    % }\end{minipage}
\end{table*}

    % \begin{minipage}{0.35\linewidth}{

\begin{table}[t]
	\centering
	% \footnotesize
	% \renewcommand{\arraystretch}{0.9}
	\resizebox{0.99\columnwidth}{!}{
	\begin{tabular}{cc|cc|cc|cc}
    \multicolumn{1}{c|}{\multirow{2}{*}{Pretrain}} & 
    \multicolumn{1}{c|}{\multirow{2}{*}{Model}} & 
    \multicolumn{2}{c|}{ADNI} &
    \multicolumn{2}{c|}{OASIS-3} & 
    \multicolumn{2}{c}{ABIDE}
    \\ \cline{3-8}
    \multicolumn{1}{c|}{} & \multicolumn{1}{c|}{} & \multicolumn{1}{c}{Ma-F1} & \multicolumn{1}{c|}{Mi-F1} & \multicolumn{1}{c}{Ma-F1} & \multicolumn{1}{c|}{Mi-F1} & \multicolumn{1}{c}{Ma-F1} & \multicolumn{1}{c}{Mi-F1}
    \\ \hline \hline
    % Jigsaw
    \multicolumn{1}{c|}{\multirow{6}{*}{Jigsaw}}   & \multirow{2}{*}{MLP} &
    \multicolumn{1}{c}{0.551} & 0.742 &
    \multicolumn{1}{c}{0.201} & 0.643 &
    \multicolumn{1}{c}{0.652} & 0.724
    \\ 
    \multicolumn{1}{c|}{} & &
    \multicolumn{1}{c}{\scriptsize{(0.015)}} & \scriptsize{(0.021)} &
    \multicolumn{1}{c}{\scriptsize{(0.020)}} & \scriptsize{(0.018)} &
    \multicolumn{1}{c}{\scriptsize{(0.021)}} & \scriptsize{(0.023)}
    \\
    \multicolumn{1}{c|}{}   & \multirow{2}{*}{GCN} &
    \multicolumn{1}{c}{0.561} & 0.804 &
    \multicolumn{1}{c}{0.230} & 0.654 &
    \multicolumn{1}{c}{0.718} & 0.755
    \\
    \multicolumn{1}{c|}{}   & &
    \multicolumn{1}{c}{\scriptsize{(0.010)}} & \scriptsize{(0.017)} &
    \multicolumn{1}{c}{\scriptsize{(0.009)}} & \scriptsize{(0.017)} &
    \multicolumn{1}{c}{\scriptsize{(0.024)}} & \scriptsize{(0.015)} 
    \\
    \multicolumn{1}{c|}{} & \multirow{2}{*}{HetMed} & 
    \multicolumn{1}{c}{\textbf{0.761}} & \textbf{0.825} &
    \multicolumn{1}{c}{\textbf{0.233}} & \textbf{0.687} &
    \multicolumn{1}{c}{\textbf{0.827}} & \textbf{0.831}
    \\ 
    \multicolumn{1}{c|}{}   & &
    \multicolumn{1}{c}{\scriptsize{(0.025)}} & \scriptsize{(0.010)} &
    \multicolumn{1}{c}{\scriptsize{(0.015)}} & \scriptsize{(0.016)} &
    \multicolumn{1}{c}{\scriptsize{(0.009)}} & \scriptsize{(0.006)}
    \\ \cline{1-8}
    % Rotation
    \multicolumn{1}{c|}{\multirow{6}{*}{Rotation}} & \multirow{2}{*}{MLP} & 
    \multicolumn{1}{c}{0.529} & 0.700 &
    \multicolumn{1}{c}{0.200} & 0.635 &
    \multicolumn{1}{c}{0.669} & 0.712 
    \\ 
    \multicolumn{1}{c|}{} & &
    \multicolumn{1}{c}{\scriptsize{(0.017)}} & \scriptsize{(0.021)} &
    \multicolumn{1}{c}{\scriptsize{(0.015)}} & \scriptsize{(0.020)} &
    \multicolumn{1}{c}{\scriptsize{(0.021)}} & \scriptsize{(0.017)}
    \\
    \multicolumn{1}{c|}{} & \multirow{2}{*}{GCN} & 
    \multicolumn{1}{c}{0.541} & 0.709 &
    \multicolumn{1}{c}{0.216} & 0.667 &
    \multicolumn{1}{c}{0.708} & 0.741
    \\
    \multicolumn{1}{c|}{}   & &
    \multicolumn{1}{c}{\scriptsize{(0.014)}} & \scriptsize{(0.020)} &
    \multicolumn{1}{c}{\scriptsize{(0.011)}} & \scriptsize{(0.017)} &
    \multicolumn{1}{c}{\scriptsize{(0.007)}} & \scriptsize{(0.010)}
    \\
    \multicolumn{1}{c|}{}& \multirow{2}{*}{HetMed} & 
    \multicolumn{1}{c}{\textbf{0.632}} & \textbf{0.796} &
    \multicolumn{1}{c}{\textbf{0.220}} & \textbf{0.675} &
    \multicolumn{1}{c}{\textbf{0.807}} & \textbf{0.817}
    \\ 
    \multicolumn{1}{c|}{}&  & 
    \multicolumn{1}{c}{\scriptsize{(0.018)}} & \scriptsize{(0.010)} &
    \multicolumn{1}{c}{\scriptsize{(0.019)}} & \scriptsize{(0.014)} &
    \multicolumn{1}{c}{\scriptsize{(0.006)}} & \scriptsize{(0.004)}
    \\ \cline{1-8}
    % Exemplar
    \multicolumn{1}{c|}{\multirow{6}{*}{Exemplar}} & \multirow{2}{*}{MLP} &
    \multicolumn{1}{c}{0.600} & 0.810 &
    \multicolumn{1}{c}{0.226} & 0.668 &
    \multicolumn{1}{c}{0.700} & 0.744
    \\ 
    \multicolumn{1}{c|}{} & &
    \multicolumn{1}{c}{\scriptsize{(0.011)}} & \scriptsize{(0.010)} &
    \multicolumn{1}{c}{\scriptsize{(0.021)}} & \scriptsize{(0.014)} &
    \multicolumn{1}{c}{\scriptsize{(0.019)}} & \scriptsize{(0.038)}
    \\
    \multicolumn{1}{c|}{} & \multirow{2}{*}{GCN} &
    \multicolumn{1}{c}{0.654} & 0.840 &
    \multicolumn{1}{c}{\textbf{0.247}} & 0.681 &
    \multicolumn{1}{c}{0.721} & 0.755
    \\ 
    \multicolumn{1}{c|}{} & &
    \multicolumn{1}{c}{\scriptsize{(0.031)}} & \scriptsize{(0.033)} &
    \multicolumn{1}{c}{\scriptsize{(0.018)}} & \scriptsize{(0.027)} &
    \multicolumn{1}{c}{\scriptsize{(0.019)}} & \scriptsize{(0.020)}
    \\
    \multicolumn{1}{c|}{} & \multirow{2}{*}{HetMed} & 
    \multicolumn{1}{c}{\textbf{0.832}} & \textbf{0.846} &
    \multicolumn{1}{c}{\textbf{0.247}} & \textbf{0.682} &
    \multicolumn{1}{c}{\textbf{0.853}} & \textbf{0.855}
    \\
    \multicolumn{1}{c|}{} & & 
    \multicolumn{1}{c}{\scriptsize{(0.010)} } & \scriptsize{(0.012)} &
    \multicolumn{1}{c}{\scriptsize{(0.030)}} & \scriptsize{(0.021)} &
    \multicolumn{1}{c}{\scriptsize{(0.008)}} & \scriptsize{(0.005)}
    \\ \hline
    \end{tabular}
	}
	% \vspace{-1ex}
\caption{Performance on 3D medical image analysis.}
\vspace{-2ex}
\label{tab:main_result3}
\end{table}

\smallskip
\noindent \textbf{Methods Compared.}
\textbf{1) Methods for 2D medical images: }
We compare HetMed against three non graph-based feature fusion approaches, i.e., ``Feature Fusion (\textbf{Feat.}),'' ``Probability Fusion (\textbf{Prob.})'' and ``Learned Feature Fusion (\textbf{Learned})'' proposed in \citet{holste2021end}, and one graph-based approach (\textbf{Spec.})~\cite{parisot2017spectral}. Since these methods are trained in an end-to-end manner, we also train HetMed in an end-to-end manner for fair comparisons, i.e., 2D medical images are directly used as input to $f(\cdot)$ instead of pre-training $f(\cdot)$ based on non-medical described in Section 4.2.1.
Besides, to compare among the pre-training approaches, i.e., SimCLR~\cite{SimCLR} and MoCo~\cite{Moco},
% for 2D images, and Jigsaw, Rotation, and Exemplar for 3D images~\cite{taleb20203d}, 
we propose baselines that do not leverage multiplex graph structure after the concatenation of image and non-image feature, i.e., \textbf{MLP} and \textbf{GCN}.
Specifically, both MLP and GCN are trained to predict labels given a concatenated vector of image and non-image features, but GCN uses a single graph constructed based on the entire non-image features whereas MLP is solely based on the features. 
Note that the major difference between Spec. and GCN is the graph structure on which each model is applied, i.e., Spec. constructs a graph by comparing absolute values of certain features leading to an almost fully connected graph, whereas GCN constructs a graph based on the cosine similarity of given features leading to a sparse graph.
% \textcolor{red}{Note that Spec. constructs graph by comparing absolute values of certain feature solely, while GCN uses cosine similarity of given features.}
Moreover, Feat. is equivalent to MLP when the model is trained end-to-end.
\textbf{2) Methods for 3D medical images: }
Since there is no existing studies for multi-modal medical image analysis that use 3D medical images, we compare HetMed with MLP and GCN as described above. Moreover, we evaluate various pre-training strategies, i.e., Jigsaw, Rotation, and Exemplar~\cite{taleb20203d}.
Further details on compared methods are described in Appendix \ref{app: Compared Methods}.
% We primarily compare our proposed framework against three non-graph-based feature fusion approaches, i.e., ``Feature Fusion (Feat.),'' ``Probablity Fusion (Prob.)'' and ``Learned Feature Fusion (Learned.)'' proposed in \citet{holste2021end}, and one graph-based approach (Spec.)~\cite{parisot2017spectral}. 
% Since these methods are trained in an end-to-end manner, we also train our framework in an end-to-end manner, i.e., no pre-training step in described in Section 4.2, for fair comparisons.

\smallskip
\noindent \textbf{Evaluation Protocol.}
For end-to-end framework evaluation, we split the data into train/validation/test data of $60/10/30\%$ following previous work \cite{holste2021end}.
For pretraining framework evaluation, we use the whole data to pretrain the image encoder network following previous work \cite{azizi2021big}, 
% i.e., to obtain the representations of medical images, 
and split the data into train/validation/test data of $60/10/30\%$ to train the final image classifier.
We measure the performance in terms of Micro-F1 and Macro-F1 for classification. We report the test performance when the performance on validation data gives the best result.

\smallskip
\noindent \textbf{Implementation Details.}
We use ResNet-18 as our backbone image encoder $f(\cdot)$ and single layer GCN \cite{kipf2016semi} as our backbone node encoder $g_r(\cdot)$.
% More formally, the encoder $g_r(\cdot)$ architecture is defined as:
% \begin{eqnarray}
%   \mathbf{H}^{(r)}= \mathrm{GCN}^{(r)}\mathbf{(X, A)} = \sigma(\mathbf{\hat{D}}^{-1/2}\mathbf{\hat{A}}\mathbf{\hat{D}}^{-1/2}\mathbf{XW}^{(r)}) ,
% \end{eqnarray}
% where $\mathbf{H}^{(r)}$ is the type $r$ node embedding matrix, $\hat{\mathbf{A}}^{(r)} = \mathbf{A}^{(r)} + \mathbf{I}$ is the type $r$ adjacency matrix with self-loops, $\hat{\mathbf{D}}^{(r)} = \sum_{i}{\hat{\mathbf{{A}}}^{(r)}_{i}}$ is the type $r$ degree matrix, $\sigma(\cdot)$ is a nonlinear activation function such as ReLU, and $\mathbf{W}^{(r)}$ is the trainable weight matrix for type $r$ relationship.
% Further details for hyperparameters are described in Appendix 4.
For hyperparameters, 
we tune them in certain ranges as follows: learning rate $\eta$ in $\left\{0.0001, 0.0005, 0.001\right\}$, supervised loss parameter $\beta$ in $\left\{0.01, 0.1, 1.0\right\}$, node embedding dimension size $d$ in $\left\{64, 128, 256\right\}$, the number of clusters $|R|$ in $\left\{3, 4, 5\right\}$, and the graph construction threshold $\theta$ in $\left\{0.01, 0.75, 0.9\right\}$ 
for each relationship.
Further details are described in Appendix \ref{app: Implementation Details}.
% ~~~ for performance, we use five data set and various image encoder modules. Especially, for 2 dimensional image encoder, we randomly slice 3 dimensional image data set (i.e. ADNI, OASIS-3, ABIDE) to get several 2 dimensional images per person. To get one image representation from multiple 2 dimensional images for person, we use mean of multiple images (Details on Method - 2D Medical Image).

\subsection{Overall Performance}
\looseness=-1
Table~\ref{tab:main_result1} and Table~\ref{tab:main_result2} show the classification performance of the methods on the end-to-end and pretraining evaluations, respectively.
We have the following observations:
\textbf{1)} Our proposed HetMed generally performs well on all datasets compared to baseline methods not only on the proposed scheme (i.e., pretraining approach), but also end-to-end training fashion.
This verifies the benefit of considering various relationships between patients during multi-modality fusion.
% By fusing multiple modalities into multiplex network, our proposed framework could truly capture complex relationship among patients, thereby consistently outperforming other fusion methods.
\textbf{2)} We also evaluate HetMed on 3D medical images in Table \ref{tab:main_result3}. HetMed also outperforms other naive fusion methods, showing generality of the proposed framework.
\textbf{3)} It is worth noting that methods that fuse multiple modalities based on a graph structure (i.e., Spec., GCN and HetMed) perform better than naive fusion methods (i.e., Feat., Prob., Learned and MLP).
This indicates modeling the relationship between patients during the fusion helps medical decision process.
Considering that most clinical decisions in reality are made based on empirical experiences, i.e., previous similar cases of patients, it is natural to consider relationship (similarity) during the fusion process.
\textbf{4)} However, among the graph-based methods, HetMed performs the best. This indicates that there exist multiple types of features that should be considered during medical decision process and also during multi-modality fusion process.
% \textbf{5)} Comparing Table~\ref{tab:main_result1} and Table~\ref{tab:main_result2}, pretraining the image encoder with non-medical image data based on recently proposed self-supervised contrastive methods (i.e., SimCLR and MoCo) helps medical image analysis as argued in \citet{azizi2021big}, which has been overlooked in previous fusion methods.
\textbf{5)} Comparing Table~\ref{tab:main_result1} and Table~\ref{tab:main_result2}, pretraining the image encoder with non-medical image data helps medical image analysis as argued in \citet{azizi2021big}, which has been overlooked in previous fusion methods.

\subsection{Model Analysis}

\begin{figure}[t]
    \centering
    \includegraphics[width=0.95\columnwidth]{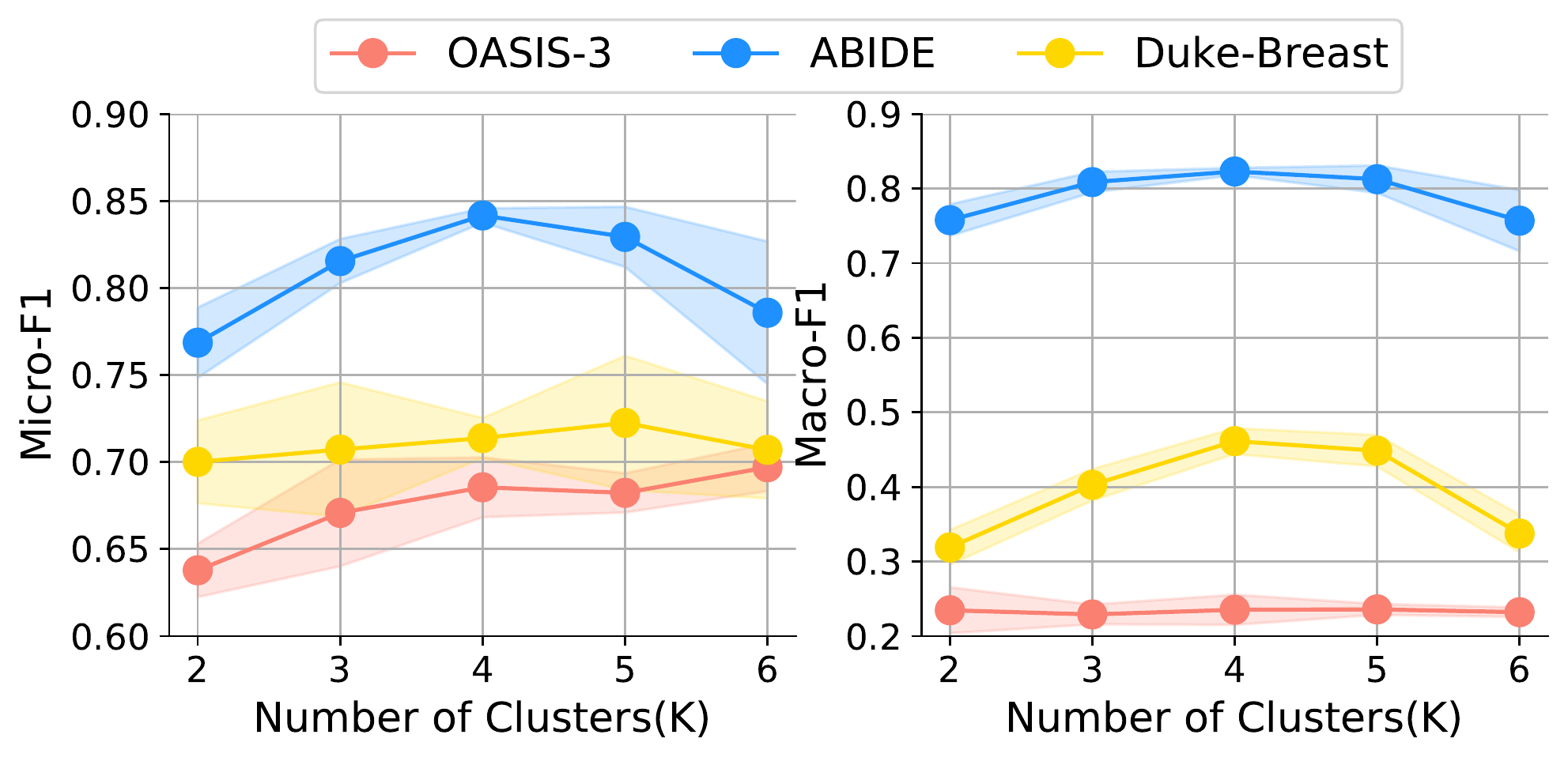}
    \captionof{figure}{Effect of number of clusters $|R|$.}
    \label{fig: Cluster Robustness}
    % }\end{minipage}
    % \vspace{-2ex}
\end{figure}

\noindent \textbf{Number of Clusters.} 
Figure \ref{fig: Cluster Robustness} shows the sensitivity analysis on the hyperparameter $|R|$ of HetMed. 
Since $|R|$ is the number of relation types between patients, it determines how complex the relationship between patients is to be modeled.
Note that HetMed becomes equivalent to a single graph framework (i.e., GCN in Table \ref{tab:main_result2} and \ref{tab:main_result3}) if $|R|$ equals to 1.
We observe that $|R| = 4$ generally gives the best performance.
On the other hand, too few or many clusters deteriorate performance of HetMed.
When $|R|$ is small, it lacks capability to model the complex relationship between patients.
When $|R|$ is large, it can get larger than the number of relation types inherent in the data, which leads to redundant information between multiple types of relationship.
Furthermore, the multiplex network may include noisy relationship that is medically meaningless, thereby deteriorating the performance of HetMed.
% We analyze the sensitivity of the hyper-parameter K of our suggested frameworks. According to Figure \ref{fig: Cluster Robustness} shows that K=4 or K=5 often offers highest performance, whereas too few or too many clusters deteriorate performance of frameworks. For deteriorate performance scenario, too few clusters is almost similar with single graph frameworks (i.e. \citet{parisot2017spectral} and GCN) in perspective (). In case of too many clusters, (occurs extreme sparsity on each types of graphs.) Clearly, both of two cases cannot include patient structure.

\begin{table}[t]
\small
	\centering
	% \footnotesize
	% \renewcommand{\arraystretch}{0.8}
	% \resizebox{0.8\columnwidth}{!}{
    \begin{tabular}{c|cc|cc}
    \multicolumn{1}{c|}{\multirow{2}{*}{Model}} &  
    \multicolumn{2}{c|}{ABIDE} &
    \multicolumn{2}{c}{OASIS-3}
    \\ \cline{2-5}
    \multicolumn{1}{c|}{} & \multicolumn{1}{c}{Ma-F1} & \multicolumn{1}{c|}{Mi-F1} & \multicolumn{1}{c}{Ma-F1} & \multicolumn{1}{c}{Mi-F1} 
    \\ \hline \hline
    % K
    \multicolumn{1}{c|}{\multirow{2}{*}{Random}} &
    \multicolumn{1}{c}{0.757} & 0.766 & 
    \multicolumn{1}{c}{0.231} & 0.672
    \\ 
    \multicolumn{1}{c|}{} &
    \multicolumn{1}{c}{\scriptsize{(0.042)}} & \scriptsize{(0.041)} &
    \multicolumn{1}{c}{\scriptsize{(0.013)}} & \scriptsize{(0.018)}
    \\\hline
    \multicolumn{1}{c|}{HetMed} &
    \multicolumn{1}{c}{0.833} & 0.842 & 
    \multicolumn{1}{c}{0.235} & 0.686
    \\
    \multicolumn{1}{c|}{(Clustering-based)}   &
    \multicolumn{1}{c}{\scriptsize{(0.005)}} & \scriptsize{(0.004)} & 
    \multicolumn{1}{c}{\scriptsize{(0.007)}} & \scriptsize{(0.011)}
    \\\hline
    \multicolumn{1}{c|}{Domain} & 
    \multicolumn{1}{c}{\textbf{0.851}} & \textbf{0.853} & 
    \multicolumn{1}{c}{\textbf{0.295}} & \textbf{0.717}
    \\ 
    \multicolumn{1}{c|}{Knowledge} & 
    \multicolumn{1}{c}{\scriptsize{(0.005)}} & \scriptsize{(0.006)} & 
    \multicolumn{1}{c}{\scriptsize{(0.020)}} & \scriptsize{(0.017)}
    % \\ \cline{2-7}
    \\ \hline
\end{tabular}
% }
% \vspace{-1ex}
\caption{Performance on various feature splitting strategies.}
\vspace{-1ex}
\label{tab: random, domain experiment}
\end{table}

% }\end{minipage}
% \begin{minipage}{0.34\linewidth}{

\begin{figure}[t]
    \centering
    \includegraphics[width=0.95\columnwidth]{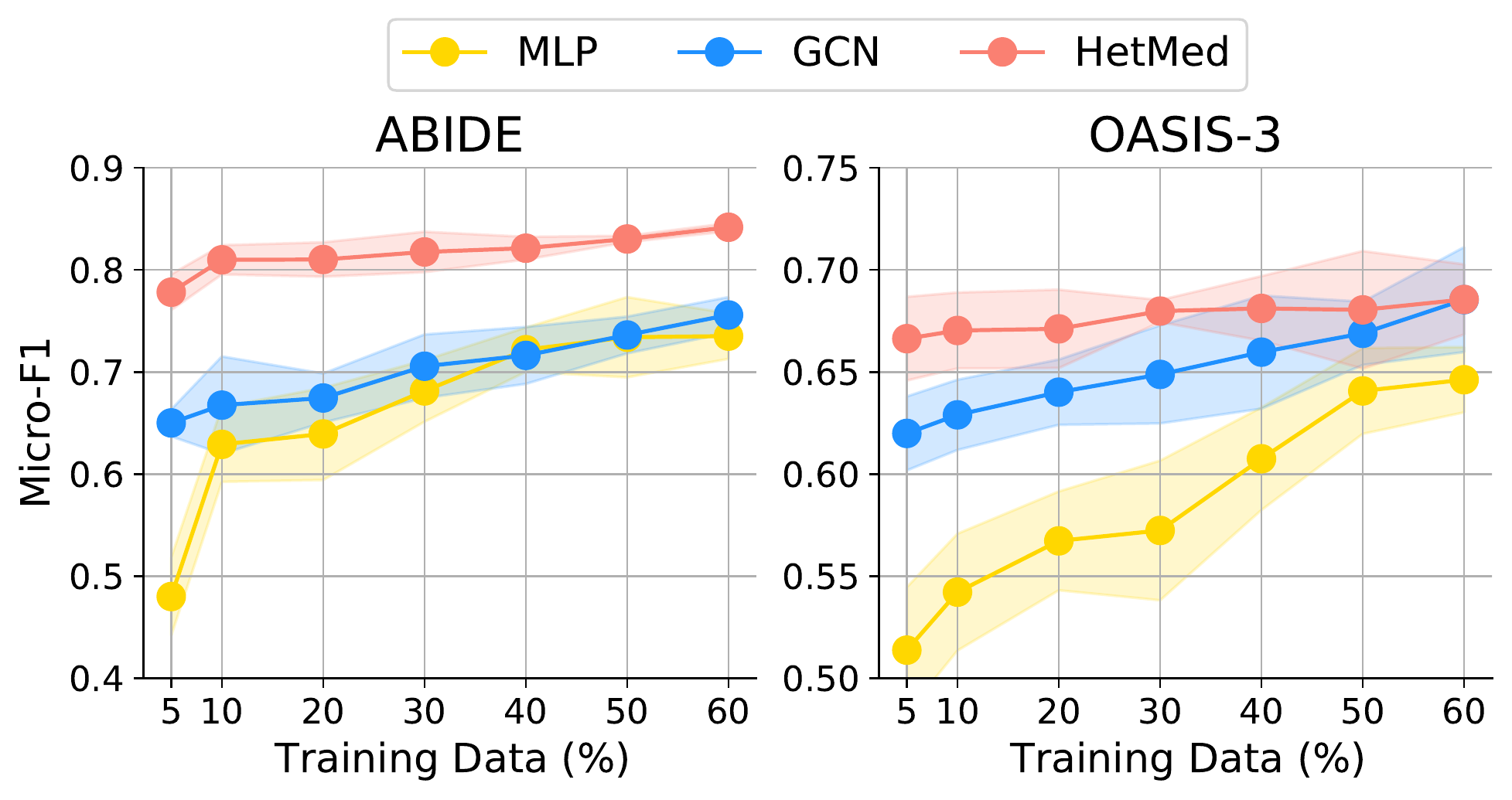}
    \captionof{figure}{Effect of number of training data.}
    \label{fig: Label Robustness}
    % }\end{minipage}
    % \vspace{-3ex}
\end{figure}

\looseness=-1
\smallskip
\noindent \textbf{Non-Image Features Splitting Strategy.}
Since a multiplex network in HetMed is constructed based on non-overlapping $|R|$ types of non-image features, it is important to have splits such that each type contains its own meaningful information.
In this regard,
% To verify the effectiveness of clustering non-image features into $|R|$ types of features, 
we compare the performance of HetMed with various feature splitting strategies in Table \ref{tab: random, domain experiment}:
% We compare the performacne with other multiplex network design techniques to ensure that K-means clustering is effective in choosing non-image types.
i) splitting features randomly, and ii) splitting features based on domain knowledge\footnote{We use clinical feature description texts of each datasets to manually split the non-image features into $|R|$ related types. Details can be found in Appendix \ref{app: Additional Experiments}.}.
% The details of domain knowledge can be found in Appendix ~\ref{app: non-image}.}.
We have the following observations:
\textbf{1)} Our proposed clustering-based splitting strategy outperforms the random splitting strategy.
We attribute this to the fact that the randomly split feature types may share similar features with one another, which hinders the construction of a meaningful multiplex network.
% randomly split subsets may contain similar types of features, thereby each relationship $\mathcal{G}^{(r)}$ contains duplicated relationship between patients.
% which indicates that the $|R|$ types of features discovered by clustering are meaningful.
\textbf{2)} Splitting non-image features based on domain knowledge outperforms the clustering-based strategy. 
% Since feature types are split based on domain knowledge rather than in a data driven way (i.e., clustering), 
We argue that feature types that are split based on domain knowledge are clinically more meaningful, which eventually leads to a multiplex network that better captures complex relationship between patients that is clinically more meaningful.
It is important to note that with some help of clinicians, the performance can be further improved, which implies that our HetMed can serve as a clinical decision support tool.
Note that splitting based on domain knowledge can be considered as an upper bound of clustering-based splitting strategy.

\smallskip
\looseness=-1
\noindent \textbf{Number of Training Data.}
% We conduct senstivity analysis on training data in Figure \ref{fig: Label Robustness}, showing the robustness of our method compared to baseline models.
Since a large volume of annotated data is rarely available in medical domain, showing robustness under the lack of labeled data is a key challenge in medical image analysis.
As shown in Figure \ref{fig: Label Robustness}, HetMed consistently produces accurate predictions even under the lack of labeled data, and the performance gap becomes larger as the number of training data gets smaller, which demonstrates the practicality of HetMed.
By modeling complex relationship into a multiplex network, HetMed becomes more robust than single relationship network (i.e., GCN).
We can also observe that graph-based methods (i.e., GCN and HetMed) are more robust than non graph-based method (i.e. MLP).
This is due to the advantage of using graph neural networks which makes decision by aggregating information from neighborhoods even with lack of label information.
Moreover, we further verify the robustness of HetMed under the number of features used in Appendix \ref{app: Additional Experiments}.

% by controlling the number of features used in Appendix 7.2. 
% eliminating certain types of features $T_r$ in Appendix 7.2.
% \textcolor{blue}{Moreover, to further verify the robustness of our proposed framework under the number of feature types, i.e., $|R|$, we conduct experiments by over various $|R|$s in Appendix 7.2.}
% medical oriented way
% In this case, since the feature types are split based on the clinical feature description texts provided in the datasets, we argue that feature types are split in a way that 
% Since domain knowledge incorporates relationship between features in the medical perspective, it can possibly capture elaborate relationship between patients. 

% In overall, the experiments indicate that incorporating multiple aspects of feature is important during feature splitting process.

\subsection{Model Practicality}
\looseness=-1
\noindent \textbf{Explainability.}
The explainability of a machine learning model is one of the most important factors in its application to the medical field.
% determining the introduction of a model in the medical field. 
Thanks to the attentive pooling mechanism in HetMed that captures the importance of each relation type, HetMed can provide explanations on which relationship has the most significant effect on the target disease.
In Figure \ref{fig: Attention Analysis}(a), we find out that model attention weights are concentrated on Type 3 relationship (i.e., cognitive abilities). 
This indicates that cognitive ability is the most important factor in determining Alzheimer's disease among the multiple clinical features.
Furthermore, in Figure \ref{fig: Attention Analysis}(b), we conduct case studies on two patients, i.e., Patient A and Patient B, with Alzheimer Disease (AD) (i.e., CDR 2), who are correctly classified as having AD by HetMed, but incorrectly classified as not having AD by GCN.
We calculate the average pairwise cosine similarity of non-image feature between patients A/B, and other patients that belong to different classes.
We find out that when computing the similarity based on all the features, patients A and B are expected to belong to CDR 0.5 (i.e., very mild impairment) and CDR 0 (i.e., normal), respectively.
However, when the similarities are computed based only on Type 3 feature discovered by HetMed to be important for the target disease, we observe that both patients A and B show the highest similarity with patients that belong to CDR 2 (i.e., moderate dementia). 
This implies that HetMed can infer the importance of each feature type, which can be used to explain the model prediction.
We further verify explainability on various datasets in Appendix \ref{app: Additional Experiments}.

% our framework can make accurate prediction on the patients with explanation based on Type 3 feature similarities, which are revealed to be the important factor of AD.

\begin{figure}[t]
    \centering
    \includegraphics[width=0.99\columnwidth]{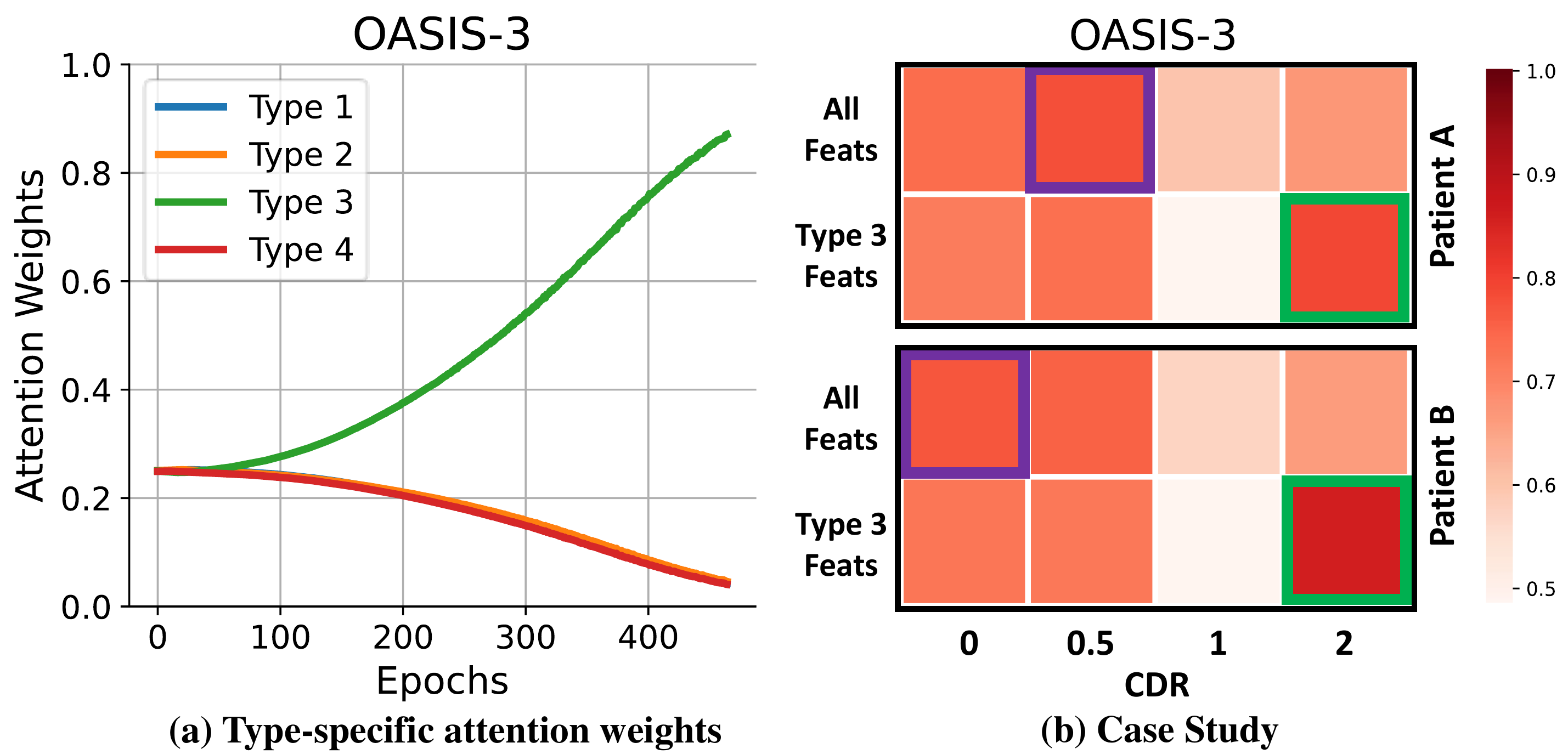}
    % \vspace{-1ex}
    \caption{Model explainability analysis.}
    % \vspace{-1ex}
    \label{fig: Attention Analysis}
\end{figure}

\smallskip
\noindent \textbf{Generalizability.}
To further verify the practicality of HetMed, we conduct experiments on the situations where new patients arrive at the hospital (Table \ref{tab: inductive experiment}). 
{We assume that new patients have provided all the required information to the hospital, which means these patients have their own medical image and non-image data. Under this situation, existing non graph-based approaches (e.g., MLP) would classify the patients based on their features alone.}
% We assume that new patients have done all the lab tests, and provided all the required information to the hospital, which means these patients have their own medical image and non-image data. Under this situation, existing non graph-based approaches (e.g., MLP) would classify the patients based on their features alone.
On the other hand, benefiting from the inductive capability of graph neural networks~\cite{hamilton2017inductive},
we propose to add the new patients into the graph that we have used for training, 
% (i.e., connect to patients that already exist in the graph), 
and classify them based on the trained model. More precisely, we first split the non-image features of the new patients into $|R|$ types of features found during training, and follow Equation~\ref{eq4} to connect them to existing patients. Having constructed the graph, we use the trained multiplex graph neural networks to obtain the embeddings for the new patients, which are then used for classification.
% Since conventional graph-based approaches require that all nodes in the graph are present during training \cite{hamilton2017inductive}, generalization ability to unseen patients have not been studied though its practicality in medical field.
For experiments, we split the data into train/validation/test data into 60/10/30\%, and use the graph that consists of patients that belong to the train split during the model training. 
We report the performance on the test data when the performance on the validation data is the best.
We observe that graph-based methods (i.e., GCN and HetMed) perform better than a non graph-based method (i.e., MLP), which demonstrates the benefit of leveraging the relationship between patients. Moreover, HetMed outperforms GCN, which again verifies that considering various relationships between patients is crucial. 
We argue that this experiment demonstrates the practicality of HetMed.
% Moreover, we combine the train and validation data for validation, and combine the train, validation, and test data for test.
% Then, we evaluate the model performance based on validation patients and test patients for validation and test, respectively.
% We observe that overall tendency was the same as previous experiments, demonstrating practicality of our proposed framework.

\begin{table}[t]
% \small
	\centering
	% \footnotesize
	% \renewcommand{\arraystretch}{0.8}
	\resizebox{0.99\columnwidth}{!}{
    \begin{tabular}{c|cc|cc|cc}
    \multicolumn{1}{c|}{\multirow{2}{*}{Model}} &  
    \multicolumn{2}{c|}{OASIS-3} &
    \multicolumn{2}{c|}{Duke-Breast} &
    \multicolumn{2}{c}{CMMD}
    \\ \cline{2-7}
    \multicolumn{1}{c|}{} & \multicolumn{1}{c}{Ma-F1} & \multicolumn{1}{c|}{Mi-F1} & \multicolumn{1}{c}{Ma-F1} & \multicolumn{1}{c|}{Mi-F1} & \multicolumn{1}{c}{Ma-F1} & \multicolumn{1}{c}{Mi-F1} 
    \\ \hline \hline
    % K
    \multicolumn{1}{c|}{\multirow{2}{*}{MLP}} &
    \multicolumn{1}{c}{0.197} & 0.646 & 
    \multicolumn{1}{c}{0.359} & 0.622 & 
    \multicolumn{1}{c}{0.484} & 0.642
    \\ 
    \multicolumn{1}{c|}{} &
    \multicolumn{1}{c}{\scriptsize{(0.017)}} & \scriptsize{(0.029)} &
    \multicolumn{1}{c}{\scriptsize{(0.011)}} & \scriptsize{(0.023)} &
    \multicolumn{1}{c}{\scriptsize{(0.028)}} & \scriptsize{(0.053)}
    \\
    \multicolumn{1}{c|}{\multirow{2}{*}{GCN}} &
    \multicolumn{1}{c}{0.203} & 0.674 & 
    \multicolumn{1}{c}{0.360} & 0.653 & 
    \multicolumn{1}{c}{0.592} & 0.670
    \\
    \multicolumn{1}{c|}{}   &
    \multicolumn{1}{c}{\scriptsize{(0.016)}} & \scriptsize{(0.015)} & 
    \multicolumn{1}{c}{\scriptsize{(0.023)}} & \scriptsize{(0.006)} & 
    \multicolumn{1}{c}{\scriptsize{(0.018)}} & \scriptsize{(0.027)}
    \\
    \multicolumn{1}{c|}{\multirow{2}{*}{HetMed}} & 
    \multicolumn{1}{c}{\textbf{0.217}} & \textbf{0.684} & 
    \multicolumn{1}{c}{\textbf{0.362}} & \textbf{0.742} & 
    \multicolumn{1}{c}{\textbf{0.631}} & \textbf{0.748}
    \\ 
    \multicolumn{1}{c|}{} & 
    \multicolumn{1}{c}{\scriptsize{(0.009)}} & \scriptsize{(0.012)} & 
    \multicolumn{1}{c}{\scriptsize{(0.012)}} & \scriptsize{(0.032)} & 
    \multicolumn{1}{c}{\scriptsize{(0.014)}} & \scriptsize{(0.013)}
    % \\ \cline{2-7}
    \\ \hline
\end{tabular}}
% \vspace{-1ex}
\caption{Performance on generalization.}
% \vspace{-3ex}
\label{tab: inductive experiment}
\end{table}

% \vspace{-1ex}
\section{Conclusion}
\looseness=-1
In this paper, we propose a general framework called \textbf{HetMed} for fusing multiple modalities of medical data, which provides heterogeneous and complementary information on a single patient.
Instead of naively fusing medical data, we propose to fuse multiple modalities into a multiplex network that contains complex relational information between patients.
By doing so, the proposed framework HetMed captures important information for clinical decision by considering various aspects of the given data.
Through experiments on the variety of multi-modal medical data, we empirically show the effectiveness of HetMed in fusing multiple modalities. 
A further appeal of HetMed is explainability and generalizability, which demonstrates the practicality of HetMed.
% paves the way for applying multiplex network graph neural networks to the medical field.
% showing the possibility of introduction of our framework to medical field.

\section*{Acknowledgements}
This work was supported by Institute of Information \& Communications Technology Planning \& Evaluation (IITP) grant funded by the Korean government (MSIT) (No.2022-0-00077), and the National Research Foundation of Korea(NRF) grant funded by the Korea government (MSIT) (No.2021R1C1C1009081)

\footnotesize
\bibliography{aaai23}

\clearpage
\section{Appendix}
\subsection{Datasets}
\label{app: Datasets}
To evaluate HetMed, we use three brain related datasets, i.e., Alzheimer's Disase Neuroimaging Intitiative (ADNI) \cite{petersen2010alzheimer}, Open Access Series of Imaging Studies (OASIS-3) \cite{lamontagne2019oasis} and Autism Brain Imaging Data Exchange (ABIDE) \cite{di2014autism, di2017enhancing}, and two breast-related datasets, i.e., Dynamic contrast-enhanced magnetic resonance images of breast cancer patients with tumor locations (Duke-Breast) \cite{saha2021dynamic,saha2018machine,clark2013cancer} and Chines Mammography Dataset (CMMD) \cite{cai2019breast, wang2016discrimination}.

\begin{itemize}

\item \textbf{ADNI} \cite{petersen2010alzheimer}:
Alzheimer's Disease Neuroimaging Initiative (ADNI) is a collection of multiple types of medical images (i.e., MRI, PET, DTI) and non-image clinical data (i.e., genetic information, clinical information and cognitive tests) that are related to Alzheimer's disease.
ADNI is an ongoing project that consists of different studies, i.e., ADNI 1, ADNI 2, ADNI 3, ADNI - GO. 
In our study, we use ADNI 1 dataset with T3-weighted MRI scan.
In this subset, total 417 individuals are comprised of 135 cognitive normal subjects, 203 subjects with late mild cognitive impairment, and 79 subjects with Alzheimer's disease.

\item \textbf{OASIS-3} \cite{lamontagne2019oasis}:
Open Access Series of Imaging Studies (OASIS-3) provides openly shared neuro-image, clinical and cognitive data for research purpose.
OASIS-3 comprises of 1,379 subjects with 2,842 MR sessions, 2,157 PET sessions and 1,472 CT sessions.
The participants of OASIS-3 include 755 cognitively normal and 622 individuals at various stages of cognitive decline ranging in age from 42-95yrs. 
Among the subjects, we select 979 subjects who have both modalities of medical data and have T1-weighted MRI images. 
Those 979 subjects are divided into 4 groups according to the various CDR level, i.e., 670 subjects with CDR 0, 189 subjects with CDR 0.5, 38 subjects with CDR 1 and 82 subjects with CDR 2.

\item \textbf{ABIDE} \cite{di2014autism, di2017enhancing}:
Autism Brain Imaging Data Exchange (ABIDE) dataset has been used to detect autism spectrum disorder based on brain imaging.
ABIDE involves 20 different sites, shares resting state functional magnetic resonance imaging (R-fMRI), anatomic, phenotype and clinical information of 1,112 subjects.
We select 977 subjects in same pipeline of MRI scans processing. The subjects comprised 473 individuals with autism spectrum disorder and 504 healthy controls individuals.

\item \textbf{Duke-Breast} \cite{saha2021dynamic,saha2018machine}:
Dynamic contrast-enhanced magnetic resonance images of breast cancer patients with tumor locations (Duke-Breast) has been published by The Cancer Imaging Archive (TCIA) \cite{clark2013cancer}.
It provides medical images and non-image clinical data for prediction of tumor.
Among 922 patient IDs, we select 614 IDs that consists of 44 IDs with tumor grade 1.0, 105 IDs with tumor grade 2.0, and 465 IDs with tumor grade 3.0.

\item \textbf{CMMD} \cite{cai2019breast, wang2016discrimination}:
Chinese Mammography Dataset (CMMD) has also been published by TCIA for early stage diagnosis on breast cancer. It includes 1,775 mammography studies from 1,775 patients in various Chinese institutions between 2012 and 2016. Among the patients, we select 1774 subsets that consists of 481 benign and 1293 malignant patients.

\end{itemize}

\begin{table*}
    \centering
    \small
    \begin{tabular}{c|ccccccc}
    \hline
    & Learning & Embedding & Number of & Graph & \multirow{2}{*}{$\alpha$} & \multirow{2}{*}{$\beta$} & \multirow{2}{*}{$\gamma$}\\
    & rate ($\eta$) & dim ($d$) & Clusters ($|R|$)& Thres. ($\theta$) &  &  & \\ \hline\hline
    ADNI        & 0.0005 & 256 & 4 & (0.9,0.9,0.9,0.9) & 0.001 & 0.1 & 0.0001\\
    OASIS-3     & 0.0001 & 128 & 4 & (0.75,0.75,0.9,0.9) & 0.001 & 0.1 & 0.0001\\
    ABIDE       & 0.0005 & 64 & 4 & (0.9,0.9,0.9,0.9) & 0.001 & 1.0 & 0.0001\\
    Duke-Breast  & 0.0005 & 64 & 4 & (0.75,0.9,0.75,0.75) & 0.001 & 0.01 & 0.0001\\
    CMMD        & 0.0001 & 64 & 4 & (0.9,0.9,0.9,0.75) & 0.001 & 0.01 & 0.0001\\ 
    \hline  
    \end{tabular}
\caption{Hyperparameter specifications of HetMed.}
\label{tab:hyperparameters}
\end{table*}

% \begin{table*}[t]
%     \begin{minipage}{0.55\linewidth}{
%     \centering
%     \small
%     \resizebox{1.0\linewidth}{!}{
%     \begin{tabular}{c|ccccccc}
%     \hline
%     & Learning & Embedding & Number of & Graph & \multirow{2}{*}{$\alpha$} & \multirow{2}{*}{$\beta$} & \multirow{2}{*}{$\gamma$}\\
%     & rate ($\eta$) & dim ($d$) & Clusters ($|R|$)& Thres. ($\theta$) &  &  & \\ \hline\hline
%     ADNI        & 0.0005 & 256 & 4 & (0.9,0.9,0.9,0.9) & 0.001 & 0.1 & 0.0001\\
%     OASIS-3     & 0.0001 & 128 & 4 & (0.75,0.75,0.9,0.9) & 0.001 & 0.1 & 0.0001\\
%     ABIDE       & 0.0005 & 64 & 4 & (0.9,0.9,0.9,0.9) & 0.001 & 1.0 & 0.0001\\
%     Duke-Breast  & 0.0005 & 64 & 4 & (0.75,0.9,0.75,0.75) & 0.001 & 0.01 & 0.0001\\
%     CMMD        & 0.0001 & 64 & 4 & (0.9,0.9,0.9,0.75) & 0.001 & 0.01 & 0.0001\\ 
%     \hline  
%     \end{tabular}}
%     \caption{Hyperparameter specifications of HetMed.}
%     \label{tab:hyperparameters}}\end{minipage}
%     \begin{minipage}{0.45\linewidth}{
%     \centering
%     \small
%     \resizebox{1.0\linewidth}{!}{
%     \begin{tabular}{c|c}
% 	\hline
% 		Methods & Source code \\ \hline
% 		SimCLR & \url{https://github.com/sthalles/SimCLR} \\
% 		MoCo & \url{https://github.com/facebookresearch/moco} \\
% 		3D Pre-training & \url{https://github.com/HealthML/self-supervised-3d-tasks} \\
% 		\citet{holste2021end} & \url{https://github.com/gholste/breast_mri_fusion} \\
% 		\hline
% 	\end{tabular}}
% 	\caption{Source code links of the baseline methods.}
% 	\label{tab:codes}
%     }\end{minipage}
% \end{table*}

\subsection{Compared Methods}
\label{app: Compared Methods}
In this section, we explain methods that are compared in the experiments.
\begin{itemize}
\item \textbf{Feature Fusion} \cite{holste2021end} concatenates the medical image data features obtained from a image encoder and features obtained from non-image medical data, and jointly train to produce a final prediction.
\item \textbf{Probability Fusion} \cite{holste2021end} combines the output probabilities of the independently trained models, i.e., image-only and non-image-only models, to produce a final prediction.
\item \textbf{Learned Feature Fusion} \cite{holste2021end} learns features from the image data and non-image data simultaneously, and uses the learned feature vectors to produce a final prediction.
\item \textbf{Spectral} \cite{parisot2017spectral} leverages a graph structure for fusing multiple modalities. Specifically, each node indicates a patient and each edge indicates the similarity between patients. However, this model uses dataset specific features for each node, i.e., vectorized functional connectivity matrix for ABIDE dataset and volumes of all 139 segmented brain structures for ADNI dataset, which hinders generality of the framework. In our experiments, we use extracted image features as node feature for fair comparisons.
\item \textbf{MLP} replaces Multiplex Graph Neural Networks in HetMed with a simple Multi-layer Perceptron (MLP). That is, MLP is trained to predict labels given concatenated vector of image and non-image feature.
\item \textbf{GCN} replaces Multiplex Graph Neural Networks in HetMed with a simple Graph Neural Networks (GNN)~\cite{kipf2016semi}. It only considers a single relationship between patients instead of multiple relationships as in our proposed multiplex network neural network.
\end{itemize}

% Furthermore, we compare various edge construction methods suggested in previous works.
% \begin{itemize}
% \item \textbf{Spectral}:
% \item \textbf{Fully Connected with edge weight}: 
% \item \textbf{GCN}: 
% \item \textbf{Multiplex}: 
% \end{itemize}

\subsection{Evaluation Metrics}
\label{app: Evaluation Metrics}

We evaluate the model performance in terms of Macro-F1 and Micro-F1 defined as follows:
% \begin{equation}
%     {\text{Precision}_{k} = \frac{\mathrm{TP}_{k}}{\mathrm{TP}_{k}+\mathrm{FP}_{k}}}
% \end{equation}
% \begin{equation}
%     {\text{Recall}_{k} = \frac{\mathrm{TP}_{k}}{\mathrm{TP}_{k}+\mathrm{FN}_{k}}}
% \end{equation}
\begin{equation}
    {\text{Precision}_{macro} = \frac{\sum_{k = 1}^{K}{\text{Precision}_{k}}}{K}}
\end{equation}
\begin{equation}
    {\text{Recall}_{macro} = \frac{\sum_{k = 1}^{K}{\text{Recall}_{k}}}{K}}
\end{equation}
\begin{equation}
    {\text {Macro-F1}= 2 \times \frac{\text{Precision}_{macro} \times \text{Recall}_{macro}}{\text{Precision}_{macro} + \text{Recall}_{macro}}}
\end{equation}
\begin{equation}
    {\text{Precision}_{micro} = \frac{\sum_{k = 1}^{K}{\text{TP}_{k}}}{\sum_{k = 1}^{K}{\text{TP}_{k} + \text{FP}_{k}}}}
\end{equation}
\begin{equation}
    {\text{Recall}_{micro} = \frac{\sum_{k = 1}^{K}{\text{TP}_{k}}}{\sum_{k = 1}^{K}{\text{TP}_{k} + \text{FN}_{k}}}}
\end{equation}
\begin{equation}
    {\text {Micro-F1}= 2 \times \frac{\text{Precision}_{micro} \times \text{Recall}_{micro}}{\text{Precision}_{micro} + \text{Recall}_{micro}}}
\end{equation}
where Precision$_k$ $= \mathrm{TP}_k / (\mathrm{TP}_k+\mathrm{FP}_k)$, Recall$_k$ $= \mathrm{TP}_k / (\mathrm{TP}_k + \mathrm{FN}_k)$, TP$_k$, TN$_k$, FP$_k$, and FN$_k$ denote precision, recall and the number of true positives, true negatives, false positives, and false negatives for class $k$, respectively. Note that in mutlti-class classification where each observation has a single label, {Micro-F1} is equivalent to {Accuracy}.

\subsection{Implementation Details}
\label{app: Implementation Details}
As described in Section 5.1 of the submitted manuscript, we use ResNet-18 \cite{he2016deep} as our backbone image encoder $f(\cdot)$.
Moreover, we use GCN \cite{kipf2016semi} encoders for multiplex network training. The base encoder of HetMed is a GCN model followed by an activation function.
More formally, the architecture of the encoder $g_r(\cdot)$ is defined as:
\begin{eqnarray}
  \mathbf{H}^{(r)}= \mathrm{GCN}^{(r)}\mathbf{(X, A)} = \sigma(\mathbf{\hat{D}}^{-1/2}\mathbf{\hat{A}}\mathbf{\hat{D}}^{-1/2}\mathbf{XW}^{(r)}) ,
\end{eqnarray}
where $\mathbf{H}^{(r)}$ is the type $r$ node embedding matrix, $\hat{\mathbf{A}}^{(r)} = \mathbf{A}^{(r)} + \mathbf{I}$ is the type $r$ adjacency matrix with self-loops, $\hat{\mathbf{D}}^{(r)} = \sum_{i}{\hat{\mathbf{{A}}}^{(r)}_{i}}$ is the type $r$ degree matrix, $\sigma(\cdot)$ is a nonlinear activation function such as ReLU, and $\mathbf{W}^{(r)}$ is the trainable weight matrix for type $r$ relationship.

\smallskip
\noindent \textbf{End-to-end Framework for 2D Medical Images.}
For the experiments regarding the end-to-end framework (Table 2 in Section 5.2), we use ResNet-18 model without any pretrained weights. By directly using medical image representation $\mathbf{Z}$ and non-image feature $\mathbf{C}$ as an input of multiple network, ResNet is also being trained during whole training process.

\smallskip
\noindent \textbf{Pretraining Framework for 2D Medical Images.}
For the experiments regarding the pretraining framework (Table 3 in Section 5.2), we pretrain ResNet-50 with non-medical image dataset, i.e., SimCLR and MoCo.
Specifically, we use pretrained weights that are provided in github repos.
With the pretrained ResNet-50, we additionally train the model with medical image through MICLe loss.
Then, the weights of ResNet model is fixed during training of multiplex graph neural network.

\smallskip
\noindent \textbf{3D Medical Images.}
For the experiments regarding 3D medical images (Table 4 in Section 5.2), we follow pretraining scheme of Table 4 in Section 5.2, i.e., we first pretrain the ResNet model with 3D self-supervised medical image representation learning methods, i.e., Jigsaw, Rotation, Exemplar, and freeze the weights of the pretrained model during training multiplex graph neural network.

\smallskip
We tune the model hyperparameters in certain ranges as described in submitted manuscript. The best performing hyperparameters are reported in Table \ref{tab:hyperparameters}.

\subsection{Additional Experiments}
\label{app: Additional Experiments}

\subsubsection{Importance of Multi-Modality}
\begin{figure}[h]
    \centering
    \includegraphics[width=0.8\columnwidth]{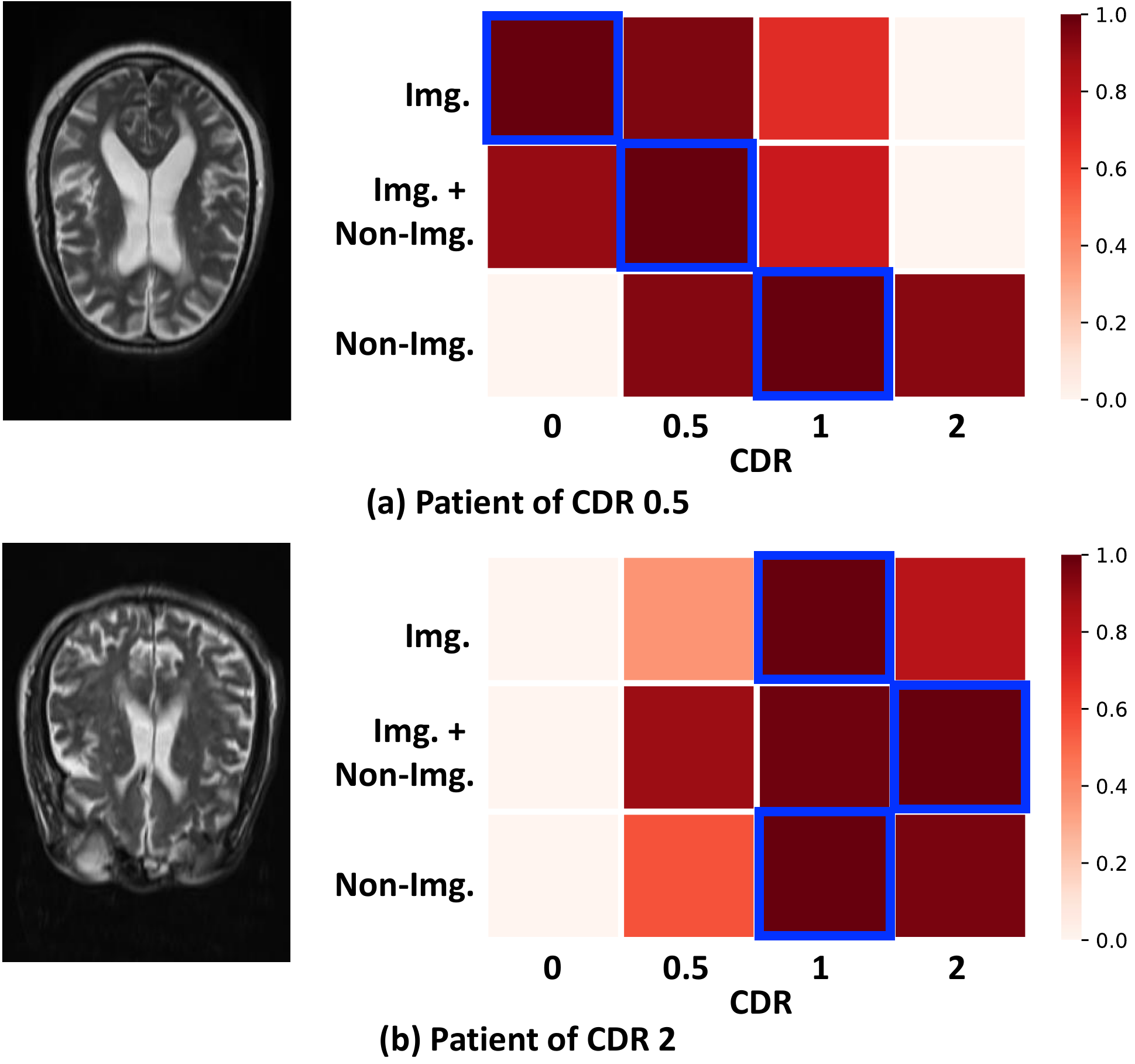}
    \caption{Case Study on OASIS-3 dataset. The element with the highest similarity value in each row is marked with a blue square.}
    \label{fig: Case Study}
\end{figure}

% \textcolor{red}{(Explain more about the experimental settings including how the features are obtained and what non-image features are used.)}
As mentioned in Section 1 of the submitted manuscript, multiple modalities of medical data provide different and complementary views of the same patient.
To corroborate our argument, we conduct case studies by calculating average pairwise cosine similarity of features between a certain patient and all other patients in various classes (Figure~\ref{fig: Case Study}).
Specifically, we calculate the average of feature similarities based on 1) only image data, 2) only non-image data, and 3) both image and non-image data.
To calculate the image-only similarity, we use the image representations produced by a pretrained image encoder.
To calculate the non-image only similarities, we use the raw non-image features given in each dataset.
To calculate the similarities based on both image and non-image data, we concatenate the image representations and non-image features, and use them for calculations.

We observe in Figure \ref{fig: Case Study} (a) that a patient with CDR 0.5 is shown to be similar to patients with CDR 0 when using only medical image data. On the other hand, the patient is shown to be similar to CDR 1 when using only non-image data. This shows that multiple modalities of medical data provide different views both of which may be incorrect.
% Those modalities provide different but both faulty clinical decision on the same patient.
That is, in this case, if the model is based only on the image data, a proactive medical decision would be dangerous since the model turns out to underestimate the situation.
% is not aware of the seriousness of the situation 
On the other hand, if the model is based only on the non-image data, 
the patient might suffer from various side effects (e.g., dizziness, insomnia, headache, etc.) since the model overestimates the situation which leads to over-treatments.
However, by incorporating both modalities (i.e. Img.+Non-Img. in Figure \ref{fig: Case Study} (a)), the model provides patients with accurate clinical decisions, which facilitates a proactive medical decision and a prevention of over-treatments.
Figure \ref{fig: Case Study} (b) shows another case study in which considering both modalities at the same time can only provide a correct clinical decision.
% We can also observe the case where both modalities offer incorrect clinical decision, while incorporating the modalities can make accurate decision in Figure \ref{fig: Case Study} (b).

\begin{table*}[t]
    \begin{minipage}{0.5\linewidth}{
    \centering
    \includegraphics[width=0.75\columnwidth]{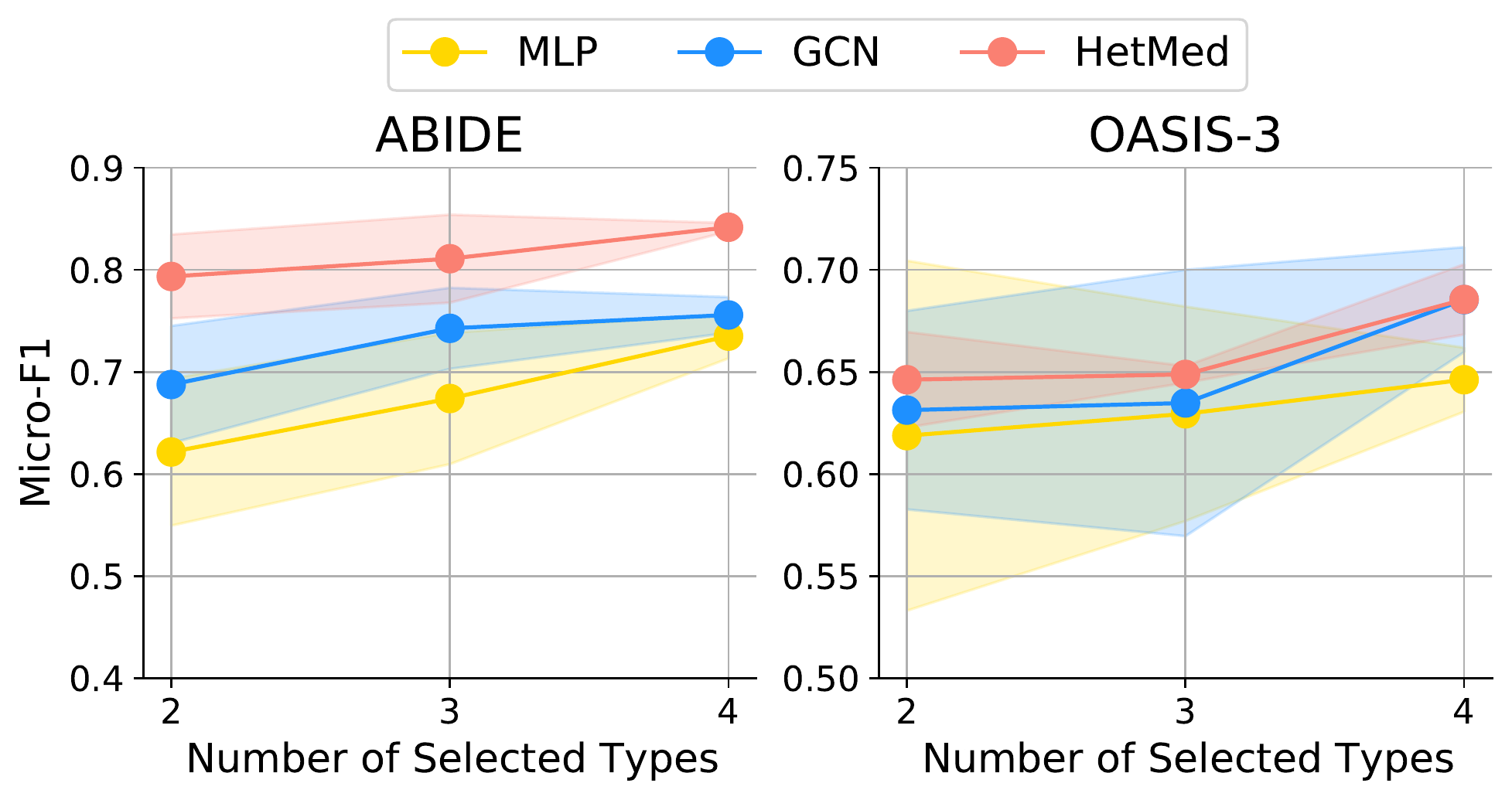}
    \captionof{figure}{Effect of number of non-image features.}
    \label{fig: Feature Robustness}
    }\end{minipage}
    \begin{minipage}{0.5\linewidth}{
    \centering
    \small
    \resizebox{0.8\linewidth}{!}{
    \begin{tabular}{c|c|cc|cc}
    \multicolumn{1}{c|}{\multirow{2}{*}{$|R|$}} &  
    \multicolumn{1}{c|}{\multirow{2}{*}{Method}} &  
    \multicolumn{2}{c|}{OASIS-3} &
    \multicolumn{2}{c}{ABIDE}
    \\ \cline{3-6}
    \multicolumn{1}{c|}{} &
    \multicolumn{1}{c|}{} & \multicolumn{1}{c}{Macro} & \multicolumn{1}{c|}{Micro} & \multicolumn{1}{c}{Macro} & \multicolumn{1}{c}{Micro} 
    \\ \hline \hline
    % K
    \multicolumn{1}{c|}{\multirow{6}{*}{1}} &
    \multicolumn{1}{c|}{\multirow{2}{*}{Spec.}} &
    \multicolumn{1}{c}{0.220} & 0.669 &
    \multicolumn{1}{c}{0.649} & 0.776
    \\ 
    \multicolumn{1}{c|}{} &
    \multicolumn{1}{c|}{} &
    \multicolumn{1}{c}{\scriptsize{(0.211)}} & \scriptsize{(0.020)} &
    \multicolumn{1}{c}{\scriptsize{(0.010)}} & \scriptsize{(0.011)} 
    \\
    \multicolumn{1}{c|}{} &
    \multicolumn{1}{c|}{F.C.} &
    \multicolumn{1}{c}{0.202} & 0.648 &
    \multicolumn{1}{c}{0.701} & 0.717
    \\
    \multicolumn{1}{c|}{} &
    \multicolumn{1}{c|}{\scriptsize{w/ weight}} &
    \multicolumn{1}{c}{\scriptsize{(0.018)}} & \scriptsize{(0.034)} &
    \multicolumn{1}{c}{\scriptsize{(0.048)}} & \scriptsize{(0.029)} 
    \\
    \multicolumn{1}{c|}{} &
    \multicolumn{1}{c|}{\multirow{2}{*}{GCN}} &
    \multicolumn{1}{c}{\textbf{0.235}} & 0.685 &
    \multicolumn{1}{c}{0.751} & 0.756
    \\ 
    \multicolumn{1}{c|}{} &
    \multicolumn{1}{c|}{} &
    \multicolumn{1}{c}{\scriptsize{(0.025)}} & \scriptsize{(0.026)} &
    \multicolumn{1}{c}{\scriptsize{(0.016)}} & \scriptsize{(0.018)} 
    
    \\ \hline
    \multicolumn{1}{c|}{\multirow{2}{*}{4}} &
    \multicolumn{1}{c|}{\multirow{2}{*}{Multiplex}} & 
    \multicolumn{1}{c}{\textbf{0.235}} & \textbf{0.686} &
    \multicolumn{1}{c}{\textbf{0.833}} & \textbf{0.842}
    \\ 
    \multicolumn{1}{c|}{} &
    \multicolumn{1}{c|}{}   & 
    \multicolumn{1}{c}{\scriptsize{(0.020)}} & \scriptsize{(0.017)} &
    \multicolumn{1}{c}{\scriptsize{(0.005)}} & \scriptsize{(0.004)} 
    % \\ \cline{2-7}
    \\ \hline
    \end{tabular}}
    \caption{Performace on various edge construction strategies.}
    \label{tab: edge construction}
    }\end{minipage}
\end{table*}

\begin{table*}[t]
    \begin{minipage}{0.65\linewidth}{
	\centering
	\footnotesize
 	\resizebox{1.0\linewidth}{!}{
	\begin{tabular}{c|cc|cc|cc|cc|cc}
    \multicolumn{1}{c|}{\multirow{2}{*}{Model}} & 
    \multicolumn{2}{c|}{ADNI} &
    \multicolumn{2}{c|}{OASIS-3} & 
    \multicolumn{2}{c|}{ABIDE} &
    \multicolumn{2}{c|}{Duke-Breast} &
    \multicolumn{2}{c}{CMMD}
    \\ \cline{2-11}
    \multicolumn{1}{c|}{} & \multicolumn{1}{c}{Ma-F1} & \multicolumn{1}{c|}{Mi-F1} & \multicolumn{1}{c}{Ma-F1} & \multicolumn{1}{c|}{Mi-F1} & \multicolumn{1}{c}{Ma-F1} & \multicolumn{1}{c|}{Mi-F1} & \multicolumn{1}{c}{Ma-F1} &
    \multicolumn{1}{c|}{Mi-F1} & \multicolumn{1}{c}{Ma-F1} &
    \multicolumn{1}{c}{Mi-F1} 
    \\ \hline \hline
    \multirow{2}{*}{MLP} &
    \multicolumn{1}{c}{0.561} & 0.781 &
    \multicolumn{1}{c}{0.219} & 0.646 &
    \multicolumn{1}{c}{0.703} & 0.735 & 
    \multicolumn{1}{c}{0.430} & 0.652 & 
    \multicolumn{1}{c}{0.523} & 0.751
    \\
    &
    \multicolumn{1}{c} {\scriptsize{(0.022)}} & \scriptsize{(0.031)} &
    \multicolumn{1}{c} {\scriptsize{(0.017)}} & \scriptsize{(0.016)} &
    \multicolumn{1}{c} {\scriptsize{(0.023)}} & \scriptsize{(0.022)} & 
    \multicolumn{1}{c} {\scriptsize{(0.019)}} & \scriptsize{(0.019)} & 
    \multicolumn{1}{c} {\scriptsize{(0.027)}} & \scriptsize{(0.014)}
    \\
    \multirow{2}{*}{GCN} &
    \multicolumn{1}{c}{0.611} & 0.816 &
    \multicolumn{1}{c}{\textbf{0.235}} & 0.685 &
    \multicolumn{1}{c}{0.751} & 0.756 & 
    \multicolumn{1}{c}{0.440} & 0.698 & 
    \multicolumn{1}{c}{0.625} & 0.745
    \\
    &
    \multicolumn{1}{c}{\scriptsize{(0.019)}} & \scriptsize{(0.014)} &
    \multicolumn{1}{c}{\scriptsize{(0.025)}}  & \scriptsize{(0.026)} &
    \multicolumn{1}{c}{\scriptsize{(0.016)}} & \scriptsize{(0.018)} & 
    \multicolumn{1}{c}{\scriptsize{(0.017)}} & \scriptsize{(0.019)} & 
    \multicolumn{1}{c}{\scriptsize{(0.014)}} & \scriptsize{(0.022)}
    \\ \hline
    % Jigsaw
    \multirow{2}{*}{HAN} &
    \multicolumn{1}{c}{0.801} & 0.823 &
    \multicolumn{1}{c}{0.221} & 0.677 &
    \multicolumn{1}{c}{0.829} & 0.834 &
    \multicolumn{1}{c}{0.434} & 0.745 &
    \multicolumn{1}{c}{0.719} & 0.779
    \\
    &
    \multicolumn{1}{c}{\scriptsize{(0.015)}} & \scriptsize{(0.023)} &
    \multicolumn{1}{c}{\scriptsize{(0.014)}} & \scriptsize{(0.017)} &
    \multicolumn{1}{c}{\scriptsize{(0.022)}} & \scriptsize{(0.031)} &
    \multicolumn{1}{c}{\scriptsize{(0.028)}} & \scriptsize{(0.034)} &
    \multicolumn{1}{c}{\scriptsize{(0.019)}} & \scriptsize{(0.027)} 
    \\
    \multirow{2}{*}{GATNE} & 
    \multicolumn{1}{c}{0.812} & 0.820 &
    \multicolumn{1}{c}{0.226} & 0.669 &
    \multicolumn{1}{c}{0.799} & 0.802 &
    \multicolumn{1}{c}{0.423} & \textbf{0.785} &
    \multicolumn{1}{c}{0.705} & 0.767 
    \\ 
    &
    \multicolumn{1}{c}{\scriptsize{(0.018)}} & \scriptsize{(0.014)} &
    \multicolumn{1}{c}{\scriptsize{(0.008)}} & \scriptsize{(0.020)} &
    \multicolumn{1}{c}{\scriptsize{(0.023)}} & \scriptsize{(0.018)} &
    \multicolumn{1}{c}{\scriptsize{(0.010)}} & \scriptsize{(0.032)} &
    \multicolumn{1}{c}{\scriptsize{(0.019)}} & \scriptsize{(0.015)} 
    \\
    \multirow{2}{*}{DMGI} &
    \multicolumn{1}{c}{\textbf{0.851}} & \textbf{0.857} &
    \multicolumn{1}{c}{\textbf{0.235}} & \textbf{0.686} &
    \multicolumn{1}{c}{\textbf{0.833}} & \textbf{0.842} &
    \multicolumn{1}{c}{\textbf{0.447}}& 0.765 &
    \multicolumn{1}{c}{\textbf{0.720}} & \textbf{0.781}
    \\ 
    &
    \multicolumn{1}{c}{\scriptsize{(0.009)}} & \scriptsize{(0.012)} &
    \multicolumn{1}{c}{\scriptsize{(0.020)}} & \scriptsize{(0.017)} &
    \multicolumn{1}{c}{\scriptsize{(0.005)}} & \scriptsize{(0.004)} &
    \multicolumn{1}{c}{\scriptsize{(0.011)}} & \scriptsize{(0.021)} &
    \multicolumn{1}{c}{\scriptsize{(0.025)}} & \scriptsize{(0.022)}
    \\ \hline
    \end{tabular}}
    \caption{Various Multiplex Network Embedding.}
    \label{tab:multiplex_network}
    }\end{minipage}
    \begin{minipage}{0.35\linewidth}{
    \centering
    \includegraphics[width=1.0\linewidth]{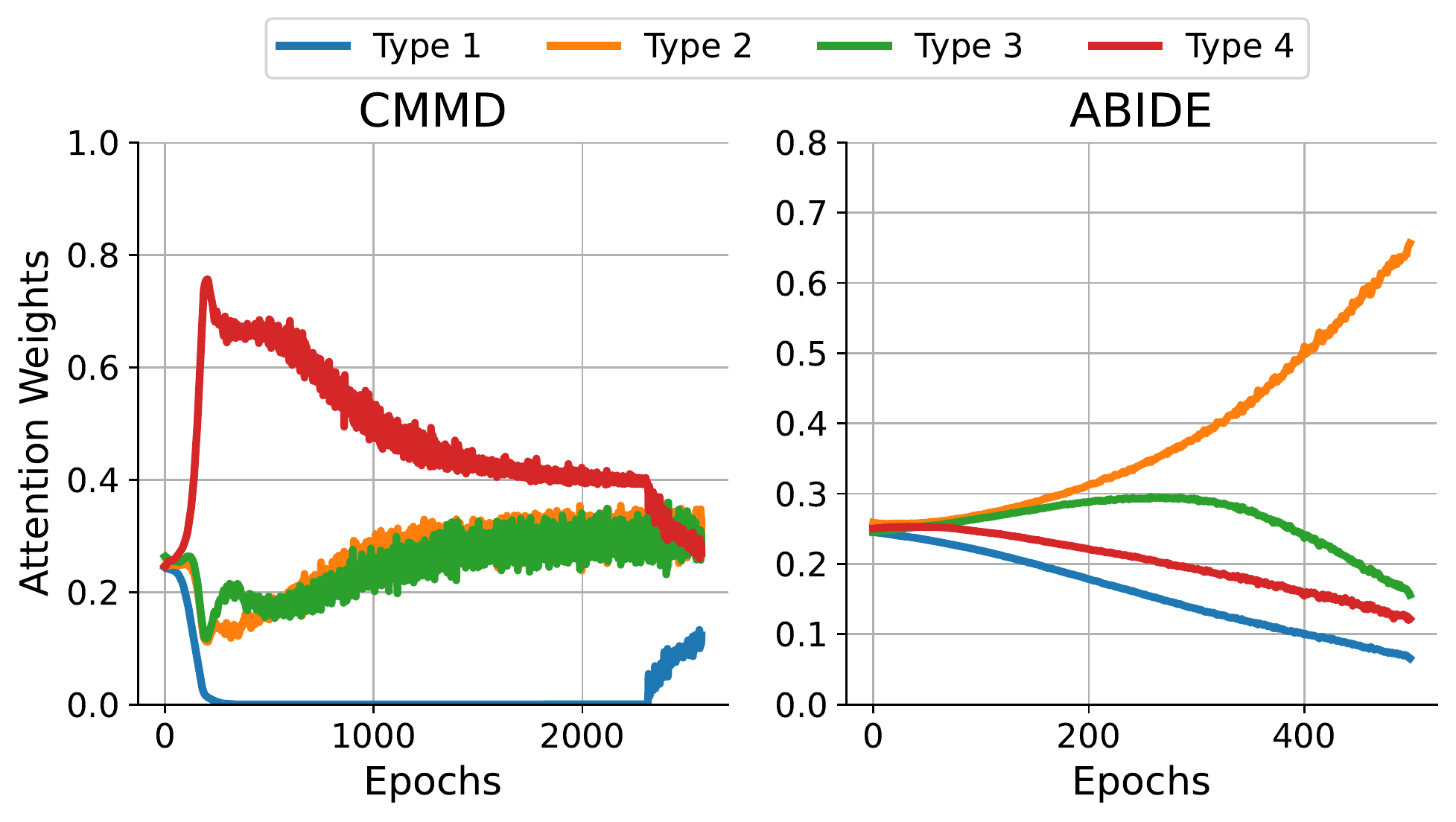}
    \captionof{figure}{Additional analyses on the learned attention weights.}
    \label{fig: Attention Analysis}
    }\end{minipage}
% 	}
\end{table*}

\subsubsection{Number of Non-Image Features}
In this section, we further verify the benefit of using multiplex network, which is shown in Section 5.3 in the submitted manuscript, i.e. robustness of multiplex network.
In this experiment, we use certain types of features during the whole experiment.
Specifically, we split non-image medical data into $|R|$ types as done before, but use only a subset among the $\mathbf{T} = \{ T_1, T_2, \cdots, T_{|R|}\}$ feature types.
By doing so, we can verify the robustness of the multiplex network-based approach under certain types of information loss. Note that we averaged over the performance of all possible combinations. For example, if only two types among four are used, we get $\binom 42=6$ combinations in total, and we average over the results for all six combinations.
As shown in Figure \ref{fig: Feature Robustness}, HetMed shows robustness even when only few feature types are used.
We argue that modeling complex relationships through a multiplex network not only benefits the model in terms of performance but also robustness.

\subsubsection{Edge Construction Strategies}
In Table \ref{tab: edge construction}, we adopt various graph construction strategies proposed in previous works \cite{parisot2017spectral, cao2021using}.
\citet{parisot2017spectral} (\textbf{Spec.}) constructs a graph by comparing absolute values of certain features leading to an almost fully connected graph, while \citet{cao2021using}  (\textbf{F.C. w/ weight}) constructs a fully connected graph with edge weights whose weights are given by the similarity between the connected patients.
Among various graph construction methods, HetMed based on a multiplex network shows the best performance, indicating the benefit of modeling complex relationship among patients.
On the other hand, we find out that constructing a single graph with cosine similarity (i.e., GCN) outperforms other single graph construction strategies.
We attribute this to the shortcomings of previous works, which connects almost every patient without regarding the informativeness of the relationships. That is, fully connected graph may inject noise to the model when aggregating the messages from connected patients.

\subsubsection{Attention Analysis}
In addition to the analysis reported in Figure 6 (a) of the submitted manuscript, 
% In Figure \ref{fig: Attention Analysis}, 
we additionally provide analyses on the attention weights learned by HetMed on other datasets used in the experiments, i.e., CMMD and ABIDE datasets, in Figure \ref{fig: Attention Analysis}.
In ABIDE dataset, we observe that the attention weights are concentrated on the Type 2 relationship, which is about stereotyped behaviors, restricted interest and social responsiveness score. 
Compared with Type 3 relationship, which is about social effect, responsiveness and resticted behavior, it is obvious that the features in the Type 2 relationship are more crucial for identifying autism, and this is automatically detected by HetMed.
% When compared to Type3 relationship (i.e. social effect and restricted behave), Type2 relationships are more significant for identifying autism. 
Furthermore, despite their fairly similar information, we find out that graph $\mathcal{G}^{(2)}$, i.e., a graph constructed by the Type 2 features, captures more informative relationship than $\mathcal{G}^{(3)}$, i.e., a graph constructed by the Type 3 features, since the social responsiveness score (i.e., SRS\_RAW\_TOTAL feature) in the Type 2 features captures more fine-grained relationship between patients than other ADOS information about social effect and responsiveness in Type 3.
% , which eventually evaluates the patients in more detail.
% Since SRS\_RAW\_TOTAL has a more specific score range (0 to 117) than other ADOS information (average: 0 to 20), Type2's graph connects related patients better than graph from Type3, even though their information is fairly similar. 
On the other hand, in CMMD dataset, all types show nearly the same attention weights when identifying breast tumor. This is because, all types of features are closely related to the diagnosis of tumor due to the lack of information.
% For this reason, Type2 (calcification of blood), Type3 (tumor associated gene expression) and Type4 (age) are closely related to the diagnosis of tumor. In order to obtain equal information from types, the model assigns nearly same attention weights to all overall types.

\subsubsection{Various Multiplex Network Embedding}
To further verify that HetMed is a model-agnostic framework, we adopt various recent multiplex network embedding methods (i.e., HAN~\cite{wang2019heterogeneous} and GATNE~\cite{cen2019representation}) into HetMed, and report results in Table \ref{tab:multiplex_network}.
We have the following observations:
\textbf{1)} We find out that HetMed with various multiplex network embedding methods outperform existing other simple fusion methods (i.e., MLP and GCN), verifying the generality of HetMed.
\textbf{2)} DMGI outperforms all other multiplex network embedding methods, indicating modeling complex relationship between patients through consensus regularization and attentive mechanism are effective.

\subsubsection{Feature Clustering Analysis}
Table \ref{tab: Feature Split} shows the features categorized based on HetMed and domain knowledge. 
In general, we observe that our proposed column-wise K-means clustering tends to categorize similar features into the same group.
For example in ADNI dataset, features that are related to simple memory tasks are grouped into Type 2 features, while features that are related to sophisticated memory tasks are grouped into Type 3 features.
Moreover, in ABIDE dataset, Type 1 features include patient specific information while Type 2 includes detailed information regarding social responsiveness and behavior scores.
We argue that by categorizing similarly characterized features into the same group, each of these groups captures distinct information, which helps to capture the complex relational information between patients.
Note that details of the features that are split based on domain knowledge reported in the experiments regarding Table 5 of the submitted manuscript can be also found in Table \ref{tab: Feature Split}.
% We noticed that the features in each type's have similar information. In the ADNI dataset, Type 2 consists of simple memory tasks for patients (e.g. memorizing digit sequences, memorize and draw simple geometric picture likes circle, square and cubic), and Type 3 consists of sophisticated memory tasks for patients (e.g. word recall task, memorize and draw picture task and naming picture task). In the ABIDE dataset, the Type 1 includes patient-specific information including sex, right- or left-handedness, and age. The Type 2 includes detailed information about social responsiveness and behavior scores, while the Type 3 includes simple score about social effect, responsiveness, and behaviors.

% \clearpage

\subsection{Pseudocode of HetMed}
    	
Algorithm~\ref{pseudocode} shows the pseudocode of HetMed.

\begin{algorithm}[h!]
		\small
        % \footnotesize
		\SetAlgoLined
		\SetKwInOut{Input}{Input}
% 		\SetKwInOut{Output}{Output}
		\Input{A input image matrix $\mathbf{B}$, A input non-image matrix $\mathbf{C}$, A label matrix $\mathbf{Y}$, Number of clusters $|R|$, Thresholds $\mathbf{\theta}$, Maximum epoch of image training $MaxEpoch_i$, Maximum epoch of fusion training $MaxEpoch_f$, Image encoder $f$, Graph encoder $g$.}
% 		\Output{The learned online representation matrix $\mathbf{H}^{\theta}\in\mathbb{R}^{N\times D}$}
		\DontPrintSemicolon
		\If{2D Image}{
            Pretrain with non-medical image
        }
		\While{not $MaxEpoch_i$}{
		    $\mathcal{L}^{\mathrm{Image}} \leftarrow \textsf{run}_{image}(\mathbf{B})$\;
		    Update \textit{f} by backpropagating $\mathcal{L}^{Image}$ 
		}
		\tcp{Learn medical img. representations}
% 		Randomly initialize the model parameters ($\xi = \theta$) \\
		Cluster $\mathbf{C}$ into $\mathbf{C}^{(r)} \leftarrow \left\{\mathbf{C}^{(1)}, \mathbf{C}^{(2)} \ldots , \mathbf{C}^{(|R|)}\right\}$\\
		\For{$r=1,2, \ldots, |R|$}{
		$\mathbf{A}^{(r)}(i, j) = \begin{cases}
                1 &\mbox{if } \rho^{(r)} (i, j) > \theta^{(r)} , \\
                0 &\mbox{otherwise, }
                \end{cases}$
		}
		$\mathbf{Z} \leftarrow f(\mathbf{B})$\;
		$\mathbf{X} \leftarrow Concatenate(\mathbf{Z},\mathbf{C})$\;
		\tcp{Multiplex network construction}
		\While{not $MaxEpoch_f$}{
		    $\mathcal{L}^{\mathrm{Fusion}} \leftarrow \textsf{run}_{fusion}(\mathbf{X},\mathbf{A}^{(1)}, \mathbf{A}^{(2)}, \ldots, \mathbf{A}^{(|R|)})$\;
		    Update $g$ by backpropagating $\mathcal{L}^{\mathrm{Fusion}}$ \;
		}
        \tcp{Train multiplex GNNs}
        \SetKwFunction{FImage}{$\textsf{run}_{image}$}
		\SetKwProg{Fn}{Function}{:}{end}
        % \SetKwProg{Fne}{Function}{:}{}

        % \SetKwProg{Fi}{Function}{:}{}

        \Fn{\FImage{$\mathbf{B}$}}{
        $\mathbf{Z} \leftarrow f(\mathbf{B})$\;
        \uIf{2D Image}{
        $\mathcal{L}^{\mathrm{Image}} \leftarrow \sum_{i=1}^{|V|} \mathcal{L}_{i}^{\mathrm{MICLe}}$
        }
        \Else{$\mathcal{L}^{\mathrm{Image}} \leftarrow \mathcal{L}^{\mathrm{3D-Image}}$}
        \KwRet $\mathcal{L}^{\mathrm{Image}}$ }

		\SetKwProg{Fn}{Function}{:}{end}
		\SetKwFunction{FMain}{$\textsf{run}_{fusion}$}

		\Fn{\FMain{$\mathbf{X},\mathbf{A}^{(1)},\mathbf{A}^{(2)},\ldots, \mathbf{A}^{(|R|)}$}}{
  
		    \For{$r=1,2, \ldots, |R|$}{
                $\mathbf{H}^{(r)}, \tilde{\mathbf{H}}^{(r)} \leftarrow g_r(\mathbf{X},\mathbf{A}^{(r)}), g_r(\tilde{\mathbf{X}}, \mathbf{A}^{(r)})$}
                
            $\mathcal{L}^{\mathrm{Fusion}} \leftarrow 
    \sum_{r \in \mathcal{R}}{\mathcal{L}^{(r)}} + \alpha \ell_{cs} + \beta \ell_{sup} + \gamma
    \| \boldsymbol{\Theta}\|^{2}$
    
			\KwRet $\mathcal{L}^{\mathrm{Fusion}}$}
		\caption{Pseudocode for HetMed.}
		\label{pseudocode}
\end{algorithm}

\begin{sidewaystable*}[b]
    \centering
	\footnotesize
	\resizebox{1.0\linewidth}{!}{
    \begin{tabular}{c|c|ll}
    \multirow{2}{*}{Dataset} & \multirow{2}{*}{Type} & \multirow{2}{*}{Ours} & Domain \\
    &  &  & Knowledge\\
    \hline \hline
    \multirow{6}{*}{ADNI}& \multirow{2}{*}{1} & \multirow{2}{*}{ADASQ4, RAVLT\_forgetting, TRABSCOR} & \multirow{2}{*}{Hippocampus, Entorhinal} \\
            &      &  &  \\ \cline{2-4}
            & \multirow{2}{*}{2} & MMSE, RAVLT\_immediate, RAVLT\_learning, LDELTOTAL, & RAVLT\_immediate, RAVLT\_learning, RAVLT\_forgetting,      \\
            &  & DIGITSCOR, mPACCdigit, mPACCtrailsB, Hippocampus, Entorhinal & RAVLT\_perc\_forgetting, LDELTOTAL, DIGITSCOR \\ \cline{2-4}
            & 3 & CDRSB, ADAS11, ADAS13, FAQ & CDRSB, ADAS13, ADAS11, TRABSCOR, FAQ, ADASQ4  \\ \cline{2-4}
            & 4 & RAVLT\_perc\_forgetting & MMSE, mPACCdigit, mPACCtrailsB  \\
    \hline
    \multirow{6}{*}{OASIS-3}& \multirow{2}{*}{1} & \multirow{2}{*}{HIS and CVD, NPI-Q, age, apoe, homehobb} & Psych Assessments, Informant Demos, \\
            &      &  & ADRC Clinical Data, Clinician Diagnosis \\ \cline{2-4}
            & \multirow{2}{*}{2} & Partcpt Family Hist., Sub Health Hist., & Phys. Neuro Findings, UPDRS, DECCLIN,  \\ 
            &      & UPDRS, ADRC Clinical Data, DECCLIN, DECIN & DECIN, FAQs, Clin. Judgements \\ \cline{2-4}
            & 3 & Psych Assessments, FAQs, Sub Demos, Clin. Judgements & Sub Health Hist., GDS, Partcpt Family Hist. \\ \cline{2-4}
            & 4 & Clinician Diagnosis, GDS, Phys. Neuro Findings, Informant Demos & age, NPI-Q, HIS and CVD, Sub Demos \\ 
    \hline
    \multirow{7}{*}{ABIDE}& 1 &  HANDEDNESS\_CATEGORY, AGE\_AT\_SCAN, SEX, FIQ, EYE\_STATUS\_AT\_SCAN & AGE\_AT\_SCAN, SEX, SRS\_RAW\_TOTAL \\ \cline{2-4}
            & \multirow{2}{*}{2} & \multirow{2}{*}{HANDEDNESS\_SCORES, ADOS\_STEREO\_BEHAV, SRS\_RAW\_TOTAL} & EYE\_STATUS\_AT\_SCAN, HANDEDNESS\_CATEGORY, \\
            &      & & ADOS\_GOTHAM\_SOCAFFECT    \\ \cline{2-4}
            & \multirow{3}{*}{3} & \multirow{3}{*}{\parbox{7.7cm}{ADOS\_GOTHAM\_SOCAFFECT, ADOS\_GOTHAM\_RRB, ADOS\_GOTHAM\_TOTAL, ADOS\_GOTHAM\_SEVERITY}} & ADOS\_GOTHAM\_SEVERITY, HANDEDNESS\_SCORES, \\ 
            &      &  & ADOS\_STEREO\_BEHAV, ADOS\_GOTHAM\_RRB, \\ 
            &      & & ADOS\_GOTHAM\_TOTAL, \\ \cline{2-4} 
            & 4 & VIQ, PIQ & VIQ, PIQ, FIQ  \\
    \hline
    \multirow{11}{*}{Duke-Breast}& \multirow{2}{*}{1} & Menopause (at diagnosis), ER, PR, Surgery, Adjuvant Radiation Therapy, & Staging(Nodes)\#(Nx replaced by -1)[N], HER2, ER, PR,      \\
            &      & Adjuvant Endocrine Therapy Medications, Pec/Chest Involvement & Staging(Metastasis)\#(Mx -replaced by -1)[M]     \\ \cline{2-4} 
            & \multirow{3}{*}{2} & HER2, Multicentric/Multifocal, Lymphadenopathy or Suspicious Nodes, & Menopause (at diagnosis), Metastatic at Presentation (Outside of Lymph Nodes)     \\ 
            &      & Definitive Surgery Type, Neoadjuvant Chemotherapy, Adjuvant Chemotherapy, & Adjuvant Chemotherapy, Adjuvant Endocrine Therapy Medications,      \\ 
            &      & Neoadjuvant Anti-Her2 Neu Therapy, Adjuvant Anti-Her2 Neu Therapy & Known Ovarian Status, Recurrence event(s)     \\ \cline{2-4}
            & \multirow{5}{*}{3}& Metastatic at Presentation (Outside of Lymph Nodes), Contralateral Breast Involvement & Surgery, Definitive Surgery Type, Neoadjuvant Radiation Therapy     \\ 
            &      & Staging(Metastasis)\#(Mx -replaced by -1)[M], Skin/Nipple Involvement, & Neoadjuvant Chemotherapy, Adjuvant Radiation Therapy       \\
            &      & Neoadjuvant Radiation Therapy, Recurrence event(s), Known Ovarian Status, & Neoadjuvant Anti-Her2 Neu Therapy, Adjuvant Anti-Her2 Neu Therapy      \\
            &      & Therapeutic or Prophylactic Oophorectomy as part of Endocrine Therapy, &  Therapeutic or Prophylactic Oophorectomy as part of Endocrine Therapy    \\
            &      & Neoadjuvant Endocrine Therapy Medications &Neoadjuvant Endocrine Therapy Medications     \\ \cline{2-4}
            & \multirow{2}{*}{4} & \multirow{2}{*}{Staging(Nodes)\#(Nx replaced by -1)[N]} & Multicentric/Multifocal, Lymphadenopathy or Suspicious Nodes  \\
            &      & &Pec/Chest Involvement, Contralateral Breast Involvement, Skin/Nipple Invovlement \\
    \hline
    \multirow{4}{*}{CMMD}& 1 & LeftRight & LeftRight \\ \cline{2-4}
            & 2 & abnormality & abnormality \\ \cline{2-4}
            & 3 & subtype & subtype \\ \cline{2-4}
            & 4 & age & age \\ 
    \hline
    \end{tabular}}
    \caption{Feature Split based on K-means clustering (Ours) and Domain Knowledge.}
    \label{tab: Feature Split}
\end{sidewaystable*}

\end{document}